\documentclass[11pt]{article}

\usepackage[preprint]{acl}

\usepackage{times}
\usepackage{latexsym}
\usepackage[table]{xcolor}

\usepackage{booktabs}
\usepackage{multirow}
\usepackage{adjustbox}
\usepackage[T1]{fontenc}
\usepackage[utf8]{inputenc}
\usepackage{microtype}
\usepackage{inconsolata}
\usepackage{inconsolata}
\usepackage[most]{tcolorbox}
\usepackage{enumitem}
\usepackage[dvipsnames]{xcolor}
\usepackage{pifont}
\usepackage{array}
\newtcolorbox[auto counter, number within=section]{prompt}[3][]{%
  enhanced,
  breakable,
  colback=#2!5!white,
  colframe=#2!75!black,
  title=\textbf{Box \thetcbcounter: #3},
  fontupper=\normalsize\fontfamily{cmss}\selectfont,
  #1
}
\definecolor{cblue}{HTML}{133844}
\definecolor{cwarmblue}{HTML}{00BDB6}
\definecolor{clightblue}{HTML}{D1F9F1}
\definecolor{ccrest}{HTML}{FFE2C8}
\definecolor{cgreen}{HTML}{DFF2EA}
\definecolor{ccherry}{HTML}{F2CAD8}
\definecolor{cviolet}{HTML}{F2ECF8}
\definecolor{cdarkviolet}{HTML}{681FB1}
\definecolor{cslate}{HTML}{546072}
\newtcolorbox{promptbox}[1][]{
  colframe=cslate,    % Frame color
  colback=white,     % Background color
  coltitle=white,            % Color of the title text
  title=#1,                  % Optional title
  rounded corners,           % Corner style
  boxrule=0.5mm,             % Frame thickness
  boxsep=5pt,                % Space between content and box
  toptitle=1mm,              % Space above the title
  bottomtitle=1mm,           % Space below the title
  left=10pt,                 % Left padding
  right=10pt,                % Right padding
  top=5pt,                   % Top padding
  bottom=5pt,                % Bottom padding
}
\newtcolorbox{promptboxcn}[1][]{
  colframe=cwarmblue,    % Frame color
  colback=white,     % Background color
  coltitle=black,            % Color of the title text
  title=#1,                  % Optional title
  rounded corners,           % Corner style
  boxrule=0.5mm,             % Frame thickness
  boxsep=5pt,                % Space between content and box
  toptitle=1mm,              % Space above the title
  bottomtitle=1mm,           % Space below the title
  left=10pt,                 % Left padding
  right=10pt,                % Right padding
  top=5pt,                   % Top padding
  bottom=5pt,                % Bottom padding
}
\newtcolorbox{promptboxgn}[1][]{
  colframe=cviolet,    % Frame color
  colback=white,     % Background color
  coltitle=black,            % Color of the title text
  title=#1,                  % Optional title
  rounded corners,           % Corner style
  boxrule=0.5mm,             % Frame thickness
  boxsep=5pt,                % Space between content and box
  toptitle=1mm,              % Space above the title
  bottomtitle=1mm,           % Space below the title
  left=10pt,                 % Left padding
  right=10pt,                % Right padding
  top=5pt,                   % Top padding
  bottom=5pt,                % Bottom padding
}
\usepackage{graphicx}
\usepackage{tikz}
\usepackage{forest}
\usetikzlibrary{trees,positioning,shapes,shadows,arrows.meta}
\usepackage{todonotes} % for comments
\usepackage{csquotes}

\usepackage{cleveref}

\title{The Rise of AfricaNLP: A Survey of Contributions, Contributors, Community Impact, and Bibliometric Analysis}

\author{
\textbf{Tadesse Destaw Belay}$^{1,7}$, 
\textbf{Kedir Yassin Hussen}$^{2}$, 
\textbf{Sukairaj Hafiz Imam}$^{3}$, \\
\textbf{Ibrahim Said Ahmad}$^{4}$, 
\textbf{Isa Inuwa-Dutse}$^{5}$,
\textbf{Abrham Belete Haile}$^{6}$,
\textbf{Grigori Sidorov}$^{1}$,\\
\textbf{Eusebio Ricardez Vázquez}$^{1}$,
\textbf{Iqra Ameer}$^{7}$,
\textbf{Idris Abdulmumin}$^{8}$, 
\textbf{Tajuddeen Gwadabe}$^{9}$,\\
\textbf{Vukosi Marivate}$^{8}$,
\textbf{Seid Muhie Yimam}$^{10}$, 
\textbf{Shamsuddeen Hassan Muhammad}$^{11}$ \\[1mm]
\footnotesize $^{1}$Instituto Politécnico Nacional, 
$^{2}$University of Gondar,
$^{3}$Northwest University Kano, 
$^{4}$Bayero University Kano,\\
\footnotesize
$^{5}$University of Huddersfield,
$^{6}$iCog Labs,
$^{7}$Pennsylvania State University,
$^{8}$University of Pretoria,\\
\footnotesize 
$^{9}$Masakhane,
$^{10}$University of Hamburg,
$^{11}$Imperial College London\
  }

\begin{document}
\maketitle
\begin{abstract}

Natural Language Processing (NLP) is undergoing constant transformation, as Large Language Models (LLMs) are driving daily breakthroughs in research and practice. In this regard, tracking the progress of NLP research and automatically analyzing the contributions of research papers provides key insights into the nature of the field and the researchers. This study explores the progress of African NLP (\textbf{AfricaNLP}) by asking (and answering) research questions about the progress of AfricaNLP (publications, NLP topics, and NLP tasks), contributions (data, method, and task), and contributors (authors, affiliated institutions, and funding bodies).
% such as: \textbf{i}) How has the nature of NLP evolved over the last two decades?, \textbf{ii}) What are the contributions of AfricaNLP papers?, and \textbf{iii}) Which individuals and organizations (authors, affiliated institutions, and funding bodies) have been involved in the development of AfricaNLP? 
We quantitatively examine two decades (2005–2025) of contributions to AfricaNLP research, using a dataset of 2.2K NLP papers, 4.9K contributing authors, and 7.8K human-annotated contribution sentences (\texttt{AfricaNLPContributions}), along with benchmark results. Our dataset and AfricaNLP research explorer tool will provide a powerful lens for tracing AfricaNLP research trends and hold potential for generating data-driven research approaches. The resource can be found in GitHub\footnote{\url{https://github.com/africanlpprogress/AfricaNLP}}. 
% The website might still be under improvement.}% AfricaNLP research reveals a winding path: significant progress has been made in developing datasets and resources for underrepresented languages, particularly following the emergence of African language–centric communities such as Masakhane. Finally, we highlight opportunities for further advancing AfricaNLP. The AfricaNLP progress tracking website\footnote{\url{https://africanlpprogress.github.io/} The website might still be under construction} is available.

\end{abstract}

\section{Introduction}

Natural Language Processing (NLP) research continues to grow at a rapid rate and has undergone significant advancements. Assessing progress, identifying challenges, and suggesting future directions can advance research in languages by showing \textit{where we are} and \textit{where we can go }\cite{adebara-2022-towards}. This need is urgent for African languages, as Africa is the most linguistically diverse continent, with over 2,000 languages, where each of them are spoken by hundreds of millions of people \cite{muhammad-etal-2023-afrisenti}.

\begin{figure}[t]
    \centering
    \includegraphics[width=\linewidth]{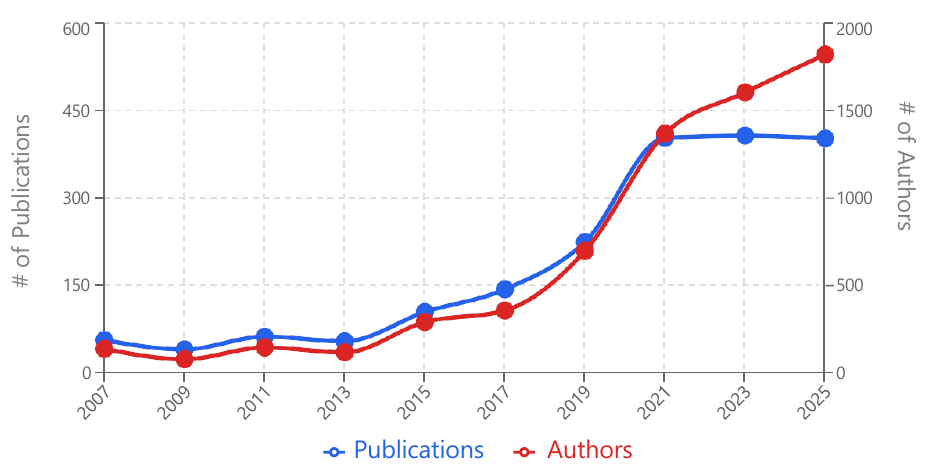}
    \caption{AfricaNLP publication history and author contributors (\textit{2005 - 2025}). Papers are keyword-based searched from Semantic Scholar and ACL events - a total of  2.2K human-verified AfricaNLP papers with more than 4.9K authors.}
    \label{fig:pub_year}
\end{figure}

Some survey work has been performed at various levels of NLP for African languages, such as domain-specific \cite{imam-etal-2025-automatic,hussen2025statelargelanguagemodels}, region/language-specific \cite{tonja-etal-2023-natural,amol2024statenlpkenyasurvey,inuwadutse2025naijanlpsurvey}, and continent-level to NLP language resources \cite{alabi2025charting}. While valuable, existing surveys are limited in their coverage of regional/country, language, modalities, and domain, and are mainly non-reproducible. We provide a comprehensive 20-year African NLP survey without any disparities with an open-source tool that tracks the progress of African NLP.

%Prior work has surveyed AfricanNLP at different scales, including language-level \cite{hausanlp}, country-level \cite{tonja-etal-2023-natural,inuwadutse2025naijanlpsurvey,amol2024statenlpkenyasurvey}, and continent-level efforts \cite{alabi2025charting}. While valuable, existing surveys are limited in their coverage across languages, regions, modalities, domains, and are often non-reproducible. In this work, we present a comprehensive 20-year survey of African NLP (2005–2025), analyzing progress, types of contributions, and contributors such as authors, affiliated institutions, and funding bodies. We also introduce the AfricaNLP Explorer, an open-source tool that tracks AfricaNLP research.

%In this work, we present a comprehensive 20-year survey of African NLP (2005–2025), analyzing the AfricaNLP progress, contribution, and contributors. We also introduced AfricaNLP Explorer, and   by an open-source and reproducible tool that systematically tracks research progress.

%without disparities in scope, region, or language.
 %We provide a comprehensive 20-year African NLP survey without any disparities with an open-source tool that tracks the progress of African NLP.

% such as domain-specific \cite{imam-etal-2025-automatic,hussen2025statelargelanguagemodels}, region or language-specific \cite{tonja-etal-2023-natural,amol2024statenlpkenyasurvey,inuwadutse2025naijanlpsurvey}, and continent-level to NLP language resources \cite{alabi2025charting}.

In NLP, contributions may take the form of new knowledge, models, datasets, or methods that have significantly improved the field \cite{khurana2023natural}. The rate of scientific literature continues to grow rapidly, and researchers face increasing challenges in keeping up with advancements in their respective fields. Thus, accurately identifying prior research and its key contributions is crucial for researchers to grasp which issues have been addressed and to effectively identify existing gaps in the literature \cite{SNYDER2019333,pramanick-etal-2025-nlpcontributions}.

In this study, we conduct a comprehensive analysis of AfricaNLP research papers published over a twenty-year span, specifically from 2005 to 2025. We seek to answer the following descriptive\footnote{Descriptive questions: which are the most basic type of quantitative research question and seeks to explain the when, where, why or how something occurred.} research questions (RQ) on AfricaNLP works: 1) How has AfricaNLP evolved over the past two decades in terms of research focus and publication trends? 2) What types of contributions (knowledge, methods, datasets) are most prevalent in AfricaNLP publications?  3) Who are the main contributors (authors, institutions, and supporting organizations) to AfricaNLP? 4) Where are the AfricaNLP papers affiliated, inside or outside the continent? 5) How does the number of language speakers correlate with the availability of language resources/publications? %This work aims to address the aforementioned research questions. % as well as other AfricaNLP-related questions.
% \todo[inline]{Can we rewrite the RQ 1 for clarity? Can we remove the Bracket in RQ 3? "This research aims to address the aforementioned research questions as well as other related questions."}

~\\The main contributions of this study are summarized as follows:
\begin{itemize}[leftmargin=3mm]
      \item We present the first comprehensive, two-decade survey of AfricaNLP research (2005–2025), analyzing publications, authors, affiliations, supporters, NLP topics, and tasks (§\ref{sec:progress}).

    \item We introduce \texttt{AfricaNLPContributions}, a large-scale dataset consisting of 2.2K papers, 4.9K authors, and 7.8K human-annotated contribution sentences (§\ref{sec:contributions}).

      \item We provide a detailed analysis of contributions, contributors, affiliations, and funding patterns in AfricaNLP (§\ref{sec:contributions}, §\ref{sec:contributors}).

        % \item We address more than 12 insightful descriptive research questions on the nature of AfricaNLP research (§\ref{sec:methods}, §\ref{sec:progress}, §\ref{sec:contributions}, and §\ref{sec:contributors}) 

 %, and benchmark the \texttt{AfricaNLPContributions} dataset.
    % \item We bridge t-he novel \texttt{contribution evaluation task} by benchmarking the \texttt{AfricaNLPContributions} dataset. We provide detailed analyses and insights from the results of various Pretrained Language models (PLMs) and LLMs.
\end{itemize}

\section{Related Works}
\subsection{African NLP Surveys}

Most previous surveys on AfricaNLP address domain-specific topics and studies focused on specific countries or languages. An overview of these efforts, including the languages, countries, and focus areas covered, is provided in \Cref{tab:related-work}.

% Most Previous surveys on AfricaNLP address different aspects, including domain-specific surveys, country- or language-focused surveys, and a small number of works that examine AfricaNLP progress more broadly. An overview of these efforts is provided in \Cref{tab:related-work}.

% Substantial efforts have been made in surveying various aspects of AfricaNLP research works. A summary of related survey efforts is shown in \Cref{tab:related-work}.

\begin{table*}[!h]
    \centering
    \scalebox{0.95}{
    \begin{adjustbox}{width=2.2\columnwidth, center}
        \begin{tabular}{lllp{4.0cm} p{11.0cm}}
        \toprule
        {\bf Survey paper} & {\bf \# of papers} & {\bf Years covered} & {\bf Languages/Countries} & {\bf Focus area(s)}\\
        \midrule
        \citet{azunre2021nlpghanaianlanguages} & < 20 & --- 2021& 16 Ghanaian languages &About NLPGhana organization and the state of NLP in Ghana  \\
        \citet{adebara-2022-towards} & < 50 & --- 2022&  512 African languages & African-centric approach to African languages\\               
        \citet{tonja-etal-2023-natural} & < 50 & --- 2023& 4 Ethiopian languages & NLP tools, datasets and benchmarking it  \\
        \citet{amol2024statenlpkenyasurvey} & < 50 & --- 2024& 15 Kenyan languages &NLP tools, datasets, and gaps  \\
        \citet{mussandi-wichert-2024-nlp} & < 50 & --- 2024&  30 African languages &  Identify languages that have NLP toolkit \\
        \citet{inuwadutse2025naijanlpsurvey} & 329 & --- 2025& 3 Nigerian languages &Linguistic resources and key challenges  \\
        \citet{hausanlp} & < 50 & --- 2025& 1 Nigerian languages & State of Hausa NLP (resources, LLMs, gaps)  \\
        
        \citet{imam-etal-2025-automatic} & < 20 & --- 2025&  African languages & Challenges of speech recognition for African languages  \\
        \citet{hussen2025statelargelanguagemodels} & < 50 & --- 2025&  42 African languages &  Identifies LLMs that support African languages  \\
        \citet{hu2025natural} & 54 & --- 2025&  African languages &  NLP technologies for public health in Africa\\

        \citet{alabi2025charting} & 732 & 5 (2019-2024) & 53 African Languages & Progresses of NLP tasks, language resources \\
       
        \textbf{Ours} & 1902 & 21 (2005-2025) & 517 African Languages  &African NLP contributions, Contributors, progresses, community impact\\
       
        \bottomrule
        \end{tabular}
    \end{adjustbox}
    }
    % \caption{Description of the taxonomy on NLP research contributions with examples.}
    \caption{Summary of related works. The number of papers that the survey reviewed, in the starting year (---), is unknown or includes all available works until the cut-off year mentioned, the Languages/countries the survey targeted, and the main focus area of the survey, with in ascending order of publication year.}
    \label{tab:related-work}
    % \vspace*{-4mm}
\end{table*}

~\\ \textbf{Domain-specific}
Several surveys and progress reports focus on specific domains and NLP tasks, including automatic speech recognition (ASR) for African languages \cite{imam-etal-2025-automatic}, LLMs for African languages \cite{hussen2025statelargelanguagemodels}, NLP for public health in Africa \cite{hu2025natural} and NLP tools for African languages \cite{mussandi-wichert-2024-nlp}.

~\\ \textbf{Country/Language-specific}
Community-led initiatives have also conducted surveys documenting the progress of NLP research in specific countries and languages such as EthioNLP \cite{tonja-etal-2023-natural} for the progress of Ethiopian languages, KenyaNLP \cite{amol2024statenlpkenyasurvey} in Kenyan languages, NaijaNLP \cite{inuwadutse2025naijanlpsurvey} for Nigerian languages, GhanaNLP \cite{azunre2021nlpghanaianlanguages} for Ghanaian languages and HausaNLP \cite{hausanlp} for Hausa language. 

~\\ \textbf{Africa-wide}
At AfricaNLP, \citet{adebara-2022-towards} identified key linguistic and sociopolitical challenges to developing NLP technologies for African languages and outlined potential research directions. More recently and the only continent-level effort to assess NLP progress in Africa, the work by \citet{alabi2025charting} analyzed 734 research papers on NLP for African languages published between 2019 and 2024. While valuable, their study was limited by a narrow set of search keywords (50 language names), restricted temporal coverage (5 years), and a focus on selected core NLP tasks and available datasets. This gap motivates our work.

%To date, this remains the only continent-level effort to assess NLP progress in Africa. Other surveys have been narrower in scope, concentrating on specific languages, countries, or NLP tasks, and thus do not provide a holistic view of AfricaNLP. 
 %We present the first comprehensive overview of AfricaNLP’s progress over the past two decades, highlighting key contributors, long-term trends, and the broader impact of AfricaNLP research on the global NLP community. A summary of related survey efforts is shown in \Cref{tab:related-work}. %This is for countries' policy makers, adaptation of the new AI tools.  

\subsection{NLP Contributions}
The work by \citet{d2020nlpcontributions} proposed an NLP contribution annotation scheme at the sentence and phrase level. \citet{dsouza-etal-2021-semeval} organized the first SemEval shared task on NLP contribution graph. \citet{pramanick_nature_2025} created a sentence-based NLP Contributions dataset by sampling five papers per year from ACL events between 1974 and February 2024. However, this work is limited to a specific database for the global-level contributions; where continent-level contributions, such as those from Africa, remain unexplored.
In this work, we present the first comprehensive survey of AfricaNLP’s progress over the past two decades, highlighting key contributions, contributors (authors, affiliated institutions, funding supporters, and computing resource supports), long-term trends, and the broader impact of AfricaNLP research on the global NLP community.

\section{ Survey Methodology} \label{sec:methods}

%In this section, we discuss the details of our survey methodology, including the process of searching for relevant papers using the updated PRISMA \cite{page2021prisma} survey guideline and further annotations. The illustration of our survey strategy is shown in Figure \ref{fig:anno}.
% \subsection{Searching Relevant Papers} 
 
\begin{figure*}[!t]
    \centering
    \includegraphics[width=\linewidth]{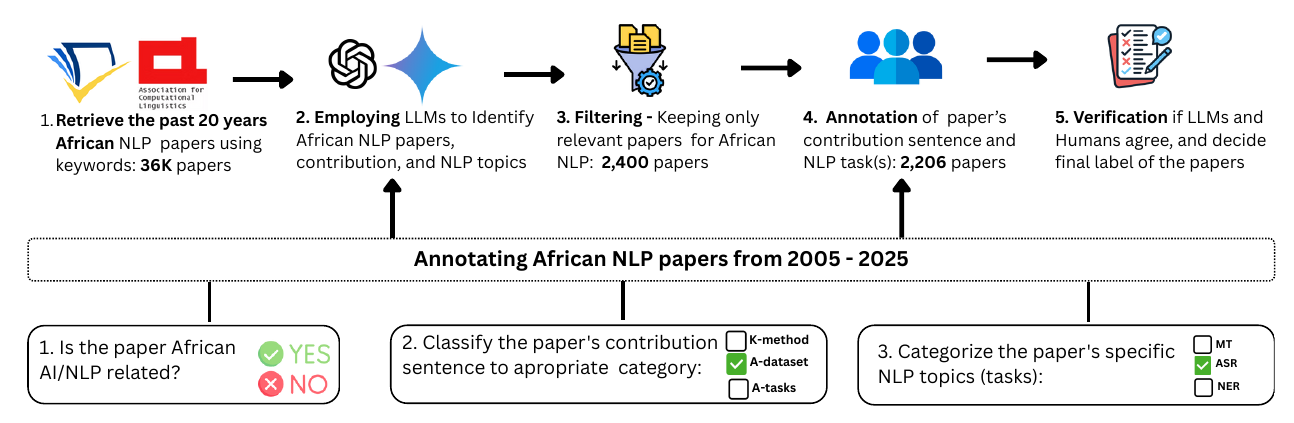}
    \caption{An illustration of the five-stage procedure for AfricaNLP paper extraction and annotation. Annotation tasks are 1) identifying AfricaNLP papers, 2) annotation of contribution sentences, and 3) categorizing the papers into the specific NLP topics (tasks).  All tasks are first annotated by a pair of LLMs (GPT-4.1 and Gemini-2.5), taking into account their agreement, and then verified/annotated by humans. The overall paper annotations of task (2) and task (3) are in a \textbf{multi-label} selection options - a paper might have more than one contribution type, labeled under more than one NLP topic and task.  The detailed annotation prompts of the LLMs are in Appendix \ref{app:prompts}.}
    \label{fig:anno}
\end{figure*}

We follow the new PRISMA framework~\cite{page2021prisma} to identify and filter papers published between 2005 and 2025. Our pipeline combines keyword-based retrieval, a multi-source approach, and LLM-assisted annotation methods. Figure~\ref{fig:anno} summarizes the overall paper retrieval and annotation process. 

~\\ \textbf{Paper Sources} We collected papers from two sources: Semantic Scholar and the ACL Anthology.
Semantic Scholar\footnote{\url{https://www.semanticscholar.org/product/api}} \cite{lo-wang-2020-s2orc} aggregates publications from more than 50 partners (including ACL, ACM, IEEE, MT Press, arXiv, dblp, and others) and provides access through its public API. %Our initial step is an exhaustive search with a matching query against the title and abstract of a paper. 
We also retrieve papers from the Association for Computational Linguistics (ACL)\footnote{\url{https://aclanthology.org/}} anthology bibliographic file, which contains prominent conference proceedings (such as EMNLP, NAACL, EACL, and ACL), journals (TACL, CL), and workshops (such as AfricaNLP).

~\\ \textbf{Searching Keywords: } To identify relevant AfricaNLP papers, we compiled a comprehensive keyword list comprising: (1) 54 African country names, (2) their adjectival forms such as African, Ethiopian, Nigerian, (3) 517 language names that are listed from the AfroLID \cite{adebara2022afrolid} work,  and (4) Grass-root African organization/community names that focus on the development of African AI/NLP, such as Masakhane, HausaNLP, EthioNLP, GhanaNLP, Lelapa, LesanAI, Lanfrica and others. The keywords co-occurrence network is presented in the Appendix, Figure \ref{fig:keyword-net}.

~\\ \textbf{Searching Criteria and Filtering:} We retrieve paper metadata using keyword search from the title and abstract with the criteria: 1) title and abstract not empty, 2) based on the Semantic Scholar topic classifier, in the field of study, there should be either \texttt{Computer Science} OR \texttt{Linguistic}, and 3) at least one keyword in the title or abstract.   We retrieved a total of 21,821 papers from the Semantic Scholar database over the past 20 years. Additionally, we explicitly retrieved 14,674 papers from ACL anthology \texttt{aclanthology.bib} file using the same set of keywords. This results in a total of 36,494 unique papers after deduplication. We retrieved the metadata of the papers, which includes the title, abstract, authors, URL, publication venue, year, citation, field of study, authors' details (name, affiliation, paper count, and h-index), and publication type (journal, conference).

\subsection{Paper Annotation}\label{sec:anno}
LLMs are increasingly applied to research support tasks, such as LLMs as a judge \cite{gu2025surveyllmasajudge}, human-in-the-loop dataset annotation \cite{calderon-reichart-2025-behalf}, and human-LLM collaborative annotation approach \cite{Wang2024human}.

As shown in Figure~\ref{fig:anno}, we used two LLMs (\texttt{GPT-4.1-mini} and \texttt{Gemini-2.5-flash}) to assist in paper annotation, specifically for identifying AfricaNLP papers, extracting contribution sentences, labeling NLP topics and tasks, and verifying contributors. A human acted as a third annotator to adjudicate the final labels. The annotation was conducted across four levels of granularity: 1) \textbf{AfricaNLP papers}: determining whether a paper is African AI/NLP-related (yes/no); papers not classified as AfricaNLP are discarded. 2) \textbf{Contribution sentences}: identifying sentences that describe contributions (knowledge and artifacts, with sub-categories described in §\ref{sec:contrib}). 3) \textbf{NLP topics and tasks}: labeling the relevant NLP topics and downstream tasks. 4) \textbf{Contributors}: extracting and verifying authors, affiliated institutions, and supporting organizations/funders from the paper. We used PyPDF2\footnote{\url{https://pypi.org/project/PyPDF2/}} to automatically extract affiliations and acknowledged bodies, with manual verification.

%1) \textbf{AfricaNLP papers}: asking \textit{Is the paper African AI/NLP-related?} to identify if it focuses (includes) on AfricaNLP or not, with yes/no options. Papers that are not AfricaNLP are discarded in this step. 2) \textbf{Contribution sentences} : identifying the contribution sentences of the paper (knowledge and artifact with its sub-categories, the details of contribution classes are shown in §\ref{sec:contrib}). 3) \textbf{The NLP topics and tasks}: labeling the specific NLP topic(s) and the downstream NLP tasks of the paper. 4) \textbf{Identifying contributors}: We extract and verify the paper contributors, affiliated institutions(organizations) and supporters/funders from the actual paper. We used PyPDF2\footnote{\url{https://pypi.org/project/PyPDF2/}} to automatically extract affiliations and acknowledged bodies, with manual verification.

For the task of identifying AfricaNLP papers, GPT-4.1 classified 1,630 papers\footnote{The overall analysis is based on paper titles and abstracts}, and Gemini classified 2,281 out of 36,494 papers, with 1,569 overlapping (69\% agreement). We manually reviewed 2,281 titles and abstracts, including Gemini-only classifications and verified 1,902 as AfricaNLP papers for further annotation. Most of the disagreements were papers in Arabic (e.g., from Morocco and Egypt) involving tasks such as sign language or optical character recognition (OCR). Examples of the paper annotation disagreement between LLMs are shown in the Appendix, Table \ref{tab:llm-disargeement}.

\subsection{Annotation Agreement}
As shown in Table \ref{tab:agreement}, the Inter-annotator Agreement (IAA) score indicates that LLMs can achieve comparable agreement scores to humans for these specific tasks. %pair with Cohen's kappa agreement between the majority vote of the three humans and the LLMs (Gemini-2.5-flash) is 82.14\%. To identify the papers, LLMs have comparable agreement with humans.
Some of the limitations of LLM-based annotations we noticed are predicting many contribution labels for a sentence and no contribution prediction is mainly shown in Gemini. This might be because we employ the lightweight and cost-effective model versions. Notably, 82\% of the papers annotated by LLMs are indeed relevant. Overall, LLMs can be used as helpful annotation assistance for NLP paper contributions and topic annotation, with a human-in-the-loop verification. 
\begin{table}[!h]
    \centering
    \resizebox{\columnwidth}{!}{
    \begin{tabular}{lccc}
    \toprule
       \textbf{Agreement}  & \textbf{AfricaNLP?} & \textbf{Contribution}& \textbf{NLP topic} \\
       \hline
       Annotators (Anno.)& 0.86 & 0.72 & 0.62\\
       Anno. Vs GPT     & 0.82 & 0.71 & 0.60\\
       Anno. Vs Gemini  & 0.82 & 0.67 & 0.61\\
       GPT Vs Gemini    & 0.85 & 0.69 & 0.69\\
       \bottomrule
    \end{tabular}}
    \caption{Inter-annotator Agreement (IAA) among three annotators and LLMs using 500 sampled papers (25 papers per/year).}
    \label{tab:agreement}
\end{table}

\begin{table*}[!t]
    \centering
    \scalebox{0.95}{
    \begin{adjustbox}{width=2.2\columnwidth, center}
        \begin{tabular}{lp{9.5cm} p{14.0cm}}
        \toprule
        {\bf Type} & {\bf Description} & {\bf Example}\\
        \midrule
        k-dataset & Describes new knowledge about datasets, such as their new properties or characteristics. & ``Our evaluation reveals a significant performance gap between high-resource languages (such as English and French) and low-resource African languages.'' --~\citet{adelani-etal-2025-irokobench} \\
        k-language & Presents new knowledge about language, such as a new property or characteristic of language. & ``When one homophone character is substituted by another, there will be a meaning change and it is against the Amharic writing regulation.'' --~\citet{belay-2021} \\

        k-method & Describes new knowledge or insights about NLP models or methods. & ``This study investigates the effectiveness of Language-Adaptive Fine-Tuning (LAFT) to improve SA performance in Hausa.'' --~\citet{sani-etal-2025-investigating} \\

        k-people & Presents new knowledge about people, humankind, society, or communities. & ``One such group, Creole languages, have long been marginalized in academic study, though their speakers could benefit from machine translation (MT).'' --~\citet{robinson-etal-2024-kreyol} \\%Nigerian Pidgin (NP) is an English-based creole language spoken by nearly 100 million people across Nigeria, and is still low-resource in NLP.
        
        k-task &Knowledge/insights about existing tasks or problem domains. & ``These results emphasize the importance of selecting appropriate pre-trained models based on linguistic considerations and task requirements.'' --~\citet{aali-etal-2024-ysp} \\

        \midrule 
        a-dataset & Introduces a new NLP dataset (i.e., textual resources such as corpora or lexicon). & ``Last, we release two Setswana LLM-translated benchmarks, MMLU-tsn and GSM8K-tsn, to measure Setswana knowledge and reasoning capabilities.'' --~\citet{brown-marivate-2025-pula} \\

        a-method & Introduces or proposes a new or novel NLP methodological approach or model to solve NLP task(s). & ``We design and implement a new MMT model framework suitable for our new generated dataset.'' --~\citet{xiao-etal-2025-text} \\
        
        a-task & Introduces or proposes a new or novel NLP task formulation (i.e., well-defined NLP problem). & ``In this work, we introduce the challenge of Meroitic decipherment as a computational task'' --~\citet{otten-anastasopoulos-2025} \\
        \bottomrule
        \end{tabular}
    \end{adjustbox}
    }
    % \caption{Description of the taxonomy on NLP research contributions with examples.}
    \caption{Description of the taxonomy for NLP research Knowledge (\textbf{k}) and artifacts (\textbf{a}) contributions with examples from the \texttt{AfricaNLPContributions} dataset.}
    \label{tab:annot_scheme}
    % \vspace*{-4mm}
\end{table*} %contribution sentence example table

% \paragraph{To what extent does this survey cover AfricaNLP research papers?}
% To ensure that as many AfricaNLP papers as possible are included in our study, we manually collect a random 50 sample research papers from various publication venues such as \texttt{Springer Nature} and contributions (model, dataset, and methods) that involve African languages, and we cross-checked to ensure that these papers are included in our collection. Among these samples, 94\% of the papers are included in our collection. The AfricaNLP papers that do not have at least a single keyword in the title/abstract from the set of searching keywords might be excluded. 

%one of the the remaining papers do not include an African country or language name in their title or abstract during searching, such as papers \citet{CVQA}, \citet{ALIYU2025100330}, and \citet{KHOBOKO2025100649} while the paper includes languages from Africa.

\section{AfricaNLP Progresses}\label{sec:progress}
\begin{figure*}[!h]
    \centering
    \includegraphics[width=\linewidth]{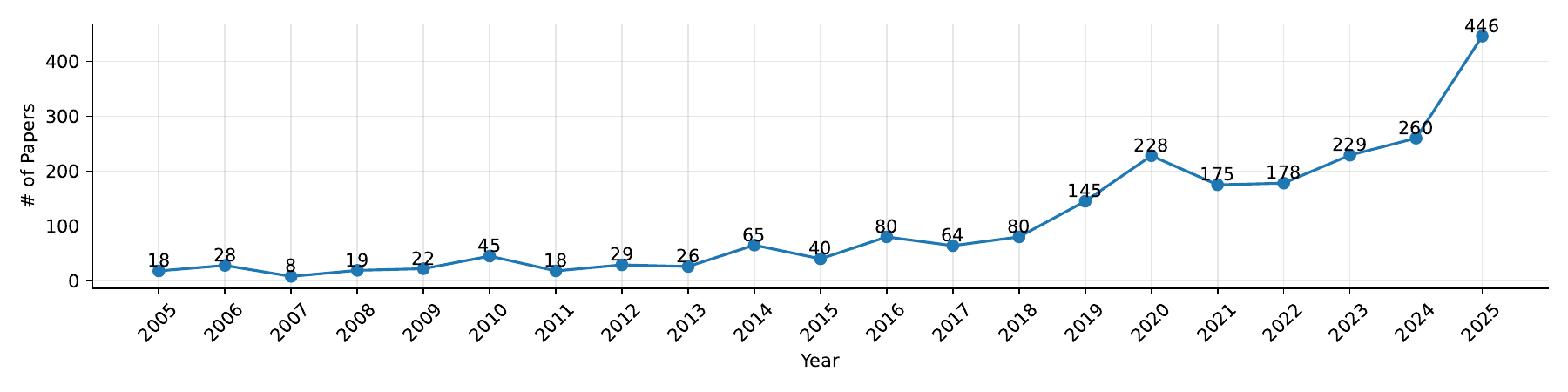}
    \caption{\textbf{The progress of AfricaNLP research over the past 20 years, 2005 – 2025}. The number of publications shows a steady increase starting from 2019.} %, with the data for the cutoff year 2025 covering to the end of July.} %AfricaNLP workshop co-located with ICLR 2020 started and has published a range of papers, including research on machine translation for 38 African languages, authored by over 35 contributors \citet{orife2020masakhanemachinetranslation}.}
    \label{fig:pub}
\end{figure*}

% \subsection{AfricaNLP Progresses}
\textbf{How has AfricaNLP research evolved over the past two decades?}
The progress of AfricaNLP over the last 20 years is shown in Figure \ref{fig:pub}. AfricaNLP has experienced steady growth since 2005, with a sharp acceleration after 2019. This surge coincides with the emergence of grassroots communities that are working on African languages. 
%Such projects includes MasakhaneMT \cite{nekoto-etal-2020-participatory}, and MasakhaNER \cite{adelani-etal-2022-masakhaner}, EthioLLM \cite{tonja-etal-2024-ethiollm}, Walia-LLM \cite{azime-etal-2024-walia}, EthioEmo \cite{belay-etal-2025-evaluating}, HausaNLP \cite{hausanlp}, and others for underrepresented African languages.
These initiatives addressed the longstanding challenge of resource scarcity through participatory, community-driven research (RQ1). In addition, the rise of AfricaNLP is due to the emergence of the transformer architecture \cite{attention-2017} and its subsequent updates, such as T5 (Text-to-Text Transfer Transformer) \cite{t5}.

In addition, the advent of the transformer architecture \cite{vaswani2017attention} and the subsequent rise of pre-trained language models (PLMs), including multilingual models such as mBERT \cite{pires2019multilingualbert}, and XLM-R \cite{conneau2020xlmr}, have further accelerated AfricaNLP progress. Their ability to handle multilingual and low-resource settings, combined with transfer learning and continual fine-tuning approaches, has enabled AfricaNLP researchers to achieve better performance in key tasks such as machine translation, named entity recognition, and sentiment analysis with limited data \cite{howard2018universal}. This is evidenced by the fact that \textit{efficient/low-resource methods for NLP} is the leading NLP topic for African languages and Keywords such as ‘pre-training’ (201 occurrences), ‘transformer’ (185), ‘fine-tuning’ (146), ‘large language models’ (145), and ‘LLM’ (141) frequently appear in paper titles and abstracts. %The synergy between the grassroots of AfricaNLP initiatives creates quality benchmark data and transformer-based pre-trained language models capabilities, which have been pivotal in driving the rapid expansion, diversification, and quality of AfricaNLP research over the past five years.
 % AfroXLM-R \cite{alabi2022afroxlmr}, and Aya \cite{ustun2024aya}

\begin{figure*}[!t]
    \centering
    \includegraphics[width=\linewidth]{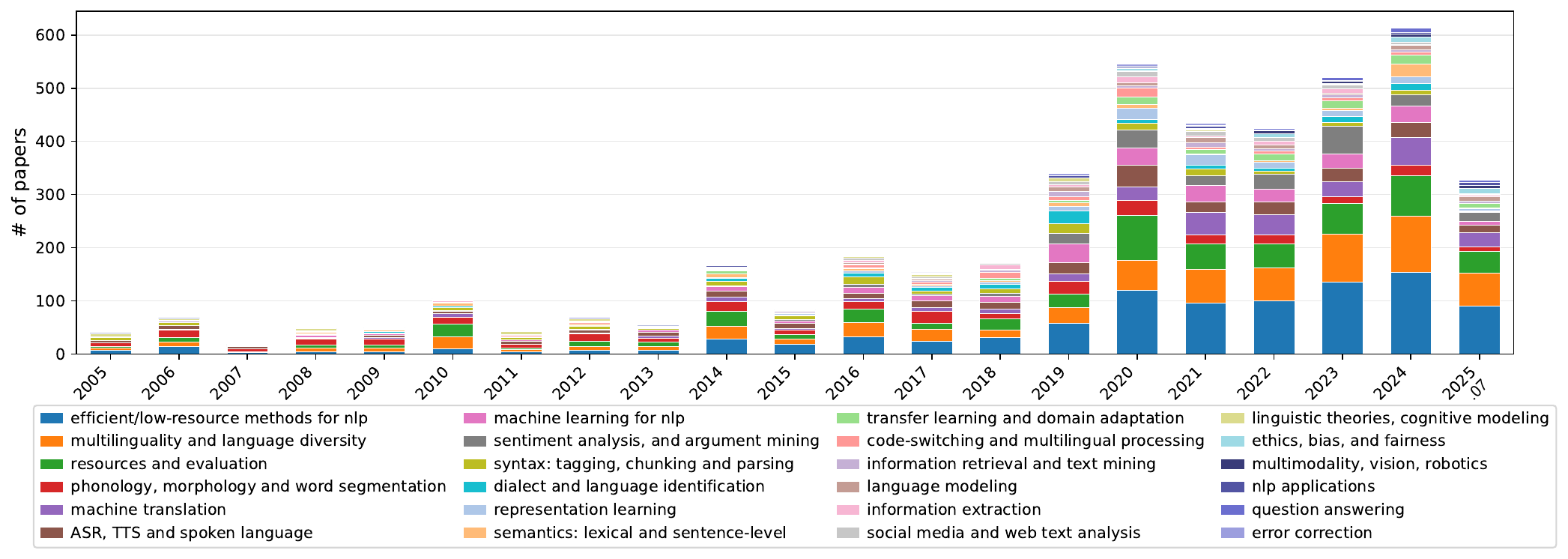}
    \caption{\textbf{AfricaNLP research increases in topic diversity},  (a paper may appear in multiple categories). The last year \texttt{2025} is the end of December. The NLP topics are sorted in descending order based on the overall total frequency of papers. Through the years, the number of topics relating to African languages and their diversity has increased from 2019 to 2025.}
    \label{fig:tasks}
\end{figure*}

~\\ \textbf{What are the main focus areas of NLP research in African languages?}
We categorize the 2.2K AfricaNLP paper under the predefined NLP topics, which encompass methods (including efficient methods for low-resource NLP), tasks (such as machine translation and speech recognition), and themes (including NLP applications, ethics, bias, and fairness). We adopt ACL Rolling Review (ARR) NLP research topics\footnote{\url{https://aclrollingreview.org/areas}}. Figure \ref{fig:tasks} shows the distribution of these NLP topics. The "\textit{efficient/low-resource methods for NLP}" NLP topic categories are the most frequently applied methods in a multilingual and linguistically diverse setup for low-resource African languages.

~\\ \textbf{How is the progress of each downstream NLP task over the years?} 
Classical tasks such as machine translation (MT), automatic speech recognition (ASR), language identification (LID), sentiment analysis, and parsing (including part-of-speech tagging and named entity recognition) are the most prevalent and widely explored tasks. \Cref{tab:nlp_tasks} in ~\Cref{app:nlptopics-tasks} presents the distribution of the top downstream NLP tasks for African languages. 

~\\ \textbf{Which NLP task(s) are getting less attention over time?} 
According to publication statistics across years for specific NLP tasks, lexicon-level NLP tasks have generally receiving limited attention. For example, Lexical Semantics had 5 papers in 2006, 15 in 2020, and 4 in 2024; Part-of-Speech Tagging had 10 papers in 2014, 25 in 2020, and 8 in 2024; Word Sense Disambiguation had 5 papers in 2022 and 3 in 2024; and Named Entity Recognition had 18 in 2023, and 5 in 2024. The reason might be that with the emergence of LLMs, they are taking over the traditional NLP tasks. With the emergence of LLMs, advanced NLP research topics have gained increasing attention. For instance, Text Generation had 14 papers in 2022, 17 in 2023, and 40 in 2024, while Language Modeling had 5 papers in 2018, 25 in 2020, and 36 in 2024. On the other hand, topics such as Reasoning (6 papers) and Vision-Language tasks (4 papers) remain relatively underexplored yet.

% ~\\  \textbf{Which NLP topic(s) and languages received attention earlier in AfricaNLP?}
% From the set of NLP research topics, 1) Syntax: tagging, chunking, and parsing \cite{fissaha-adafre-2005-part},  2) Phonology, morphology, and word segmentation \cite{hu-etal-2005-refining}, 3) ASR \cite{seid05_interspeech}, 4) OCR \cite{meshesha2007optical} are the early researched downstream NLP tasks since 2005. Languages such as Amharic, Afrikaans, Kinyarwanda, Swahili, Setswana, and Yoruba are among the earliest researched African languages since 2005. More details about the progress of specific NLP tasks are presented in Appendix \ref{app:tasks}.

\section{AfricaNLP Contributions}\label{sec:contributions}
In this section, we discuss the analysis of \texttt{AfricaNLPContributions} dataset  (§\ref{subs:contrib}) and contribution classification task experiments (§\ref{sec:contrib}).

\subsection{AfricaNLPContributions Dataset}\label{subs:contrib}

Contribution sentences are automatically annotated from the abstracts of 1902 papers with the help of LLMs and human approvals, as discussed in Section \ref{sec:anno}. 
The sentence-level contribution annotations comprised a single or a few sentences about the article’s contribution inside the abstract \cite{dsouza-etal-2021-semeval,pramanick_nature_2025}. We adopted the 8 classes NLPContributions annotation taxonomy (5 knowledge type and 3 artifacts type)  from previous work \cite{pramanick_nature_2025}, The class names are knowledge about method (\textit{k-method}), knowledge about dataset (\textit{k-dataset}), knowledge about task (\textit{k-task}), knowledge about language (\textit{k-language}), and artifacts of new methods (\textit{a-methods}), new datasets (\textit{a-datasets}), and new tasks (\textit{a-tasks}). The contribution classes are described with examples in Table \ref{tab:annot_scheme}.

\begin{figure}[h!]
    \centering
    \includegraphics[width=\linewidth]{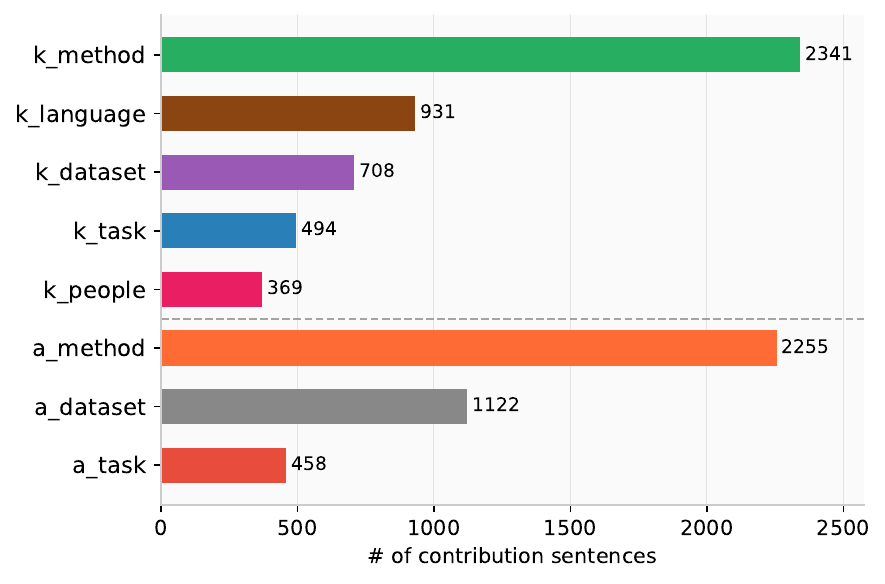}
    \caption{Contribution sentences occurrence statistics of different contribution types in the paper abstracts (a paper may have multiple contribution sentences).}%and a sentence may appear in multiple contribution categories
    \label{fig:contribution}
\end{figure}

\textbf{What kinds of contributions are made in AfricaNLP papers?} From the final 1,902 AfricaNLP papers, 7,796 unique sentences are annotated as knowledge and artifact contributions. The distributions are presented in Table \ref{tab:lbl_dist}. Among these, 27\% of the sentences are knowledge of methods (k-method), 26\% of the sentences are artifacts of methods (a-method), 12.9\% of the sentences are artifacts of the dataset (a-dataset), and 10.7\% of the sentences are knowledge of language (k-language). The least contributions are artifacts of new tasks (a-task) and knowledge about people  (k-people), 5.3\% and 4.3\% respectively. The contributions \textit{knowledge-method} and \textit{artifact-method} are the most identified collection contribution sentences in the abstract. The detailed contribution statistics across 8 labels are shown in Figure \ref{fig:contribution}.

% We analyze the occurrence of multiple contributions in a paper. From a total of class occurrences: 8,678 sentences (7,796 unique), and 882 sentences contain more than one contribution label, with most of them being knowledge of methods (k-method) and artifacts of method (a-method) together. The overall co-occurrences are shown in Figure \ref{fig:occrance}.

% \begin{figure}[h!]
%     \centering
%     \includegraphics[width=\linewidth]{images/contribution_co-ocurance.pdf}
%     \caption{Contribution co-occurrence across the 8 contribution labels, 8,678 contribution sentences with 882 (11.3\%) statements received multiple contribution labels}
%     \label{fig:occrance}
% \end{figure}

~\\ \textbf{How AfricaNLP contributions align with global NLP research?}
We compare the contributions of AfricaNLP (\texttt{AfricaNLPContributions}) with works done at global NLP (\texttt{NLPContributions}) by \citet{pramanick-etal-2025-nlpcontributions}. In the knowledge category, the dominant \texttt{NLPContributions} class is k-task, which involves creating new downstream NLP tasks. In \texttt{AfricaNLPContributions}, the focus is on k-method - developing, adopting, or evaluating NLP methods. From the artifact category, a-dataset is the dominant one in both global and AfricaNLP contributions. In artifacts, \texttt{AfricaNLPContributions} aligns with the global NLP research contribution - \texttt{NLPContributions} \cite{pramanick_nature_2025}.

\subsection{Contributions Classification Task}\label{sec:contrib}
\paragraph{Task Definition.} 
The task is to automatically classify a sentence into a predefined two broad contribution classes: \textbf{artifacts} - which encompass the development of new or novel resources (methods, tasks, or datasets) and \textbf{knowledge} - contributions that enrich the field with new insights or knowledge of language, methods, tasks, people, or datasets. Given an abstract $A$ in plaintext format, the goal was to extract a set of contribution sentences $C_{sent} = \{C_{sent_{1}}, ... ,C_{sent_{N}}\}$. We set up a multi-class classifier for the \texttt{AfricaNLPContributions} dataset to model and automatically classify a sentence into the 8 classes (5 knowledge and 3 artifact contribution classes), the details of the classes are shown in Table \ref{tab:annot_scheme}.

\paragraph{Methods.} 
We select the following pre-trained language models (PLMs) for benchmarking from different categories: BERT \cite{bert-abs-1810-04805} from general-purpose PLMs, SciBERT \cite{beltagy-etal-2019-scibert} from scientific text-based PLMs, AfroXLMR \cite{adelani-etal-2024-sib} from African-language centric, and GPT-4.1-mini \cite{gpt4.1}) from state-of-the-art LLMs.  For prompting LLM, we apply a zero-shot setting. The \texttt{AfricaNLPContributions} dataset is split into 5,847 training set, 1,170 test set, and 779 validation set, with a 75:15:10 splitting ratio, respectively.  Training details and hyperparameters are presented in Appendix \ref{app:param}.

% \begin{displayquote}
%     \textit{Compared to other regions where the AI ecosystem is shaped by big corporations or strong policies and regulation frameworks, Africa’s AI ecosystem is dominated by grassroots movements.}  -- \cite{o2024ai}
% \end{displayquote}

% \begin{table}[t]
%     \centering
%     \scalebox{1.0}{
%         \begin{tabular}{l l r r}
%         \toprule
%         {\bf Type} & {\bf Sub-typ.} & {\bf Prop. \%}& {\bf\# Sent} \\
%         \midrule
%         \multirow{5}{*}{Knowledge} & k-dataset & 8.16 & 708 \\
%         & k-language & 10.73 & 931 \\
%         & k-method & 26.98 & 2341 \\
%         & k-people & 4.25 & 369 \\
%         & k-task & 5.69 & 494 \\
%         \midrule 
%         \multirow{3}{*}{Artifact} & a-dataset & 12.93 & 1122 \\
%         & a-method & 25.99 & 2255 \\
%         & a-task & 5.28 & 458 \\
%         \bottomrule
%         \end{tabular}}
%     \caption{Distributions of \texttt{AfricaNLPContributions} classes: \textbf{Prop.} is proportion in percent and \textbf{Sent.} is number of sentences for the corresponding contribution class.}
%     \label{tab:lbl_dist}
% \end{table}

\begin{table}[t]
    \centering
\resizebox{\columnwidth}{!}{
        \begin{tabular}{l c c c c}
        \toprule
        {\bf Contrib.} & {\textbf{BERT}} & {\textbf{SciBERT}} & {\textbf{AfroXLMR}}&{\textbf{GPT*}}\\
        \midrule
        k-language & \textbf{0.69} & 0.68 & 0.65 & 0.49 \\
        k-method & 0.79 & \textbf{0.80} & 0.78 & 0.63 \\
        k-people & 0.42 & 0.49 & \textbf{0.54} & 0.49 \\
        k-task & \textbf{0.41} & 0.40 & 0.36 & 0.30 \\
        k-dataset & 0.54 & 0.51 & \textbf{0.62} & 0.55 \\

        \midrule
        a-dataset & 0.80 & 0.81 & \textbf{0.83} & 0.80 \\
        a-method & 0.84 & \textbf{0.85} & 0.84 & 0.76 \\
        a-task & 0.42 & \textbf{0.48} & 0.43 & 0.47 \\
        \midrule
        \textbf{Overall} & 0.61 & \textbf{0.63} & \textbf{0.63} & 0.50 \\        
        \bottomrule
        \end{tabular}
    }
    \caption{Performance of different models (macro F1) for contribution statement classification in a multi-label setting. GPT* is \texttt{gpt-4.1-mini}. Bold results are the best performance for the specific contribution class.}
    \label{tab:eval_result}
    % \vspace*{-5mm}
\end{table}

~\\ \textbf{Results and Discussion.}
Table \ref{tab:eval_result} shows the results. We observe that SciBERT and AfroXLMR outperform BERT and GPT, likely due to the SciBERT pre-training on a collection of scientific documents, and AfroXLMR's familiarity with African countries and language names.

% \section{Challenges in African NLP}
% \subsection{Data Scarcity}
% \begin{itemize}
%     \item The lack of large-scale, high-quality datasets is a primary bottleneck for NLP in African languages. This scarcity hampers the development of robust language models and downstream applications.
%     \item Linguistic Diversity: only 43 languages getting attention 
% \end{itemize}
%Africa is home to over 2,000 languages, many of which feature complex grammatical structures, tonal systems, and unique scripts. This diversity poses a significant challenge to NLP researchers. More than 2500 languages are listed in http://ethnologue.com/language/ and pilot languages with the number of speakers in millions

\section{AfricaNLP Contributors}\label{sec:contributors}
In this section, we provide analyses and insights to the following threefold contributors on AfricaNLP. \textbf{1) Authors}: We analyze the sets of authors (both first and co-authors) of the papers.  \textbf{2) Affiliations}: We identify affiliated institutions, organizations, and independent researchers from the authors' affiliation section of the paper.  \textbf{3) Supporters}: We identify funding, computing resource (GPU), and model API supports from the acknowledgment section of the NLP paper.
% \end{itemize}

~\\ \textbf{Who is supporting the development of AfricaNLP research?}
% From 1,902 papers, the support contributors are identified in the acknowledgment section of 412 papers. The remaining papers are either do not have an acknowledgment section or acknowledge the support of research idea development, reviewers, readers, volunteers, or data annotators. During acknowledged organization annotation, similar organization names—such as Google DeepMind and Google Research—are merged into a single organization (Google), together with support from different regional Google offices.
The National Science Foundation (NSF)\footnote{\url{https://www.nsf.gov/}} -  independent United States Federal Government Agency, Google\footnote{\url{https://www.google.org/}}, National Research Foundation of South Africa (ZA)\footnote{\url{https://www.nrf.ac.za/}}, Lacuna Fund \footnote{\url{https://lacunafund.org/}} -  an initiative co-founded by The Rockefeller Foundation, Google.org, and Canada’s International Development Research Center and others are the main supporters of the development of AfricaNLP. As noted throughout the papers, we also acknowledge that the mention of organization names here are not reflect the official policies or positions of the funding or supporting organizations. The top contributors are presented in Figure \ref{fig:org}; the complete list of supporter organizations is included in Appendix \ref{app:org}.
\begin{figure}[!h]
    \centering
    \includegraphics[width=\linewidth]{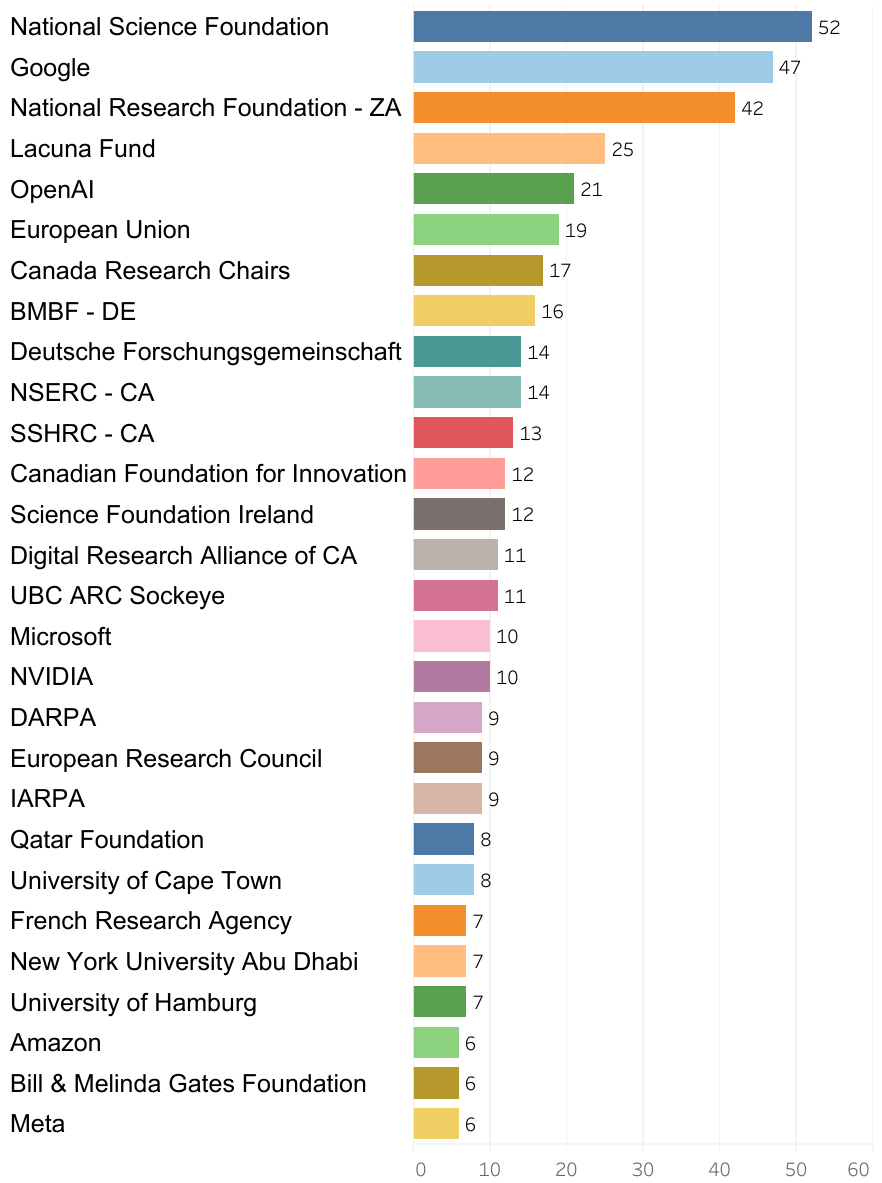}
    \caption{Top organizations with the number of publications supported, extracted from 2.2K AfricaNLP papers.}% Multiple organizations may support a work, and the entities that provide full or partial funding, computing resources such as GPUs, and entities supporting access through model API keys.
    \label{fig:org}
\end{figure}

\begin{figure}[!h]
    \centering
    \includegraphics[width=\linewidth]{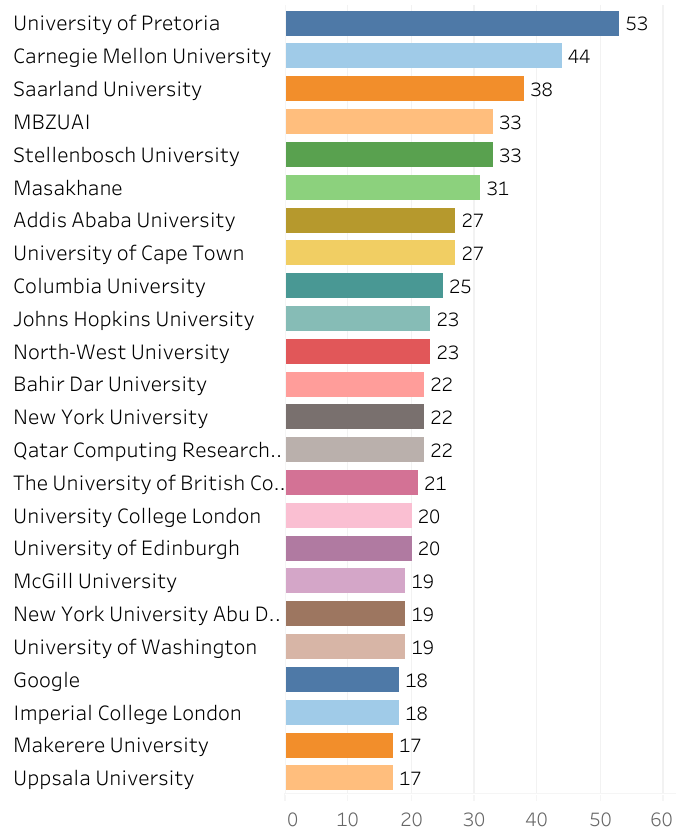}
    \caption{Top affiliated institutions/organizations/individuals of the authors (multiple institutions may be affiliated in a paper)  with the number of publications,  extracted from 2.2K AfricaNLP papers.}% Each institution/independent researcher is contributing interesting research in non-African languages or our limited set of keywords.}
    \label{fig:intitution}
\end{figure}

~\\ \textbf{Which institutions are affiliated with the development of AfricaNLP?} 
% The same approach is applied to acknowledged organizations: departments, faculties, and research labs or groups, which are grouped under the umbrella of the same organization name are recorded as the same affiliation. We extract the list of affiliated institutions from the actual papers. The affiliation institutions are identified from 1,367 papers; the other papers do not have an affiliation, either independent researchers, require a subscription to read/access the paper, or are unable ot access the PDF.  
We have identified a total of around 200 unique organizations, and Figure \ref{fig:intitution} presents the top affiliated organization. Generally, the majority of the top contributing institutions to AfricaNLP are based outside Africa, while the \textit{University of Pretoria} and \textit{Stellenbosch University} are the leaders among Africa-based affiliated institutions. The entire list of affiliated organization names is given in Appendix Table \ref{app:intitutions} and their networking in Figure \ref{fig:intitution}.%, and The full collaboration networks of affiliated institutions are shown in Figure \ref{fig:intitution}.

% \begin{figure*}[!h]
%     \centering
%     \includegraphics[width=\linewidth]{images/authors_network.pdf}
%     \caption{detail caption here}
%     \label{fig:author-net}
% \end{figure*}

%Oportunities section (sec 7)

\section{Applications and Future Work}
Advancing NLP for African languages necessitates a multifaceted approach that addresses data scarcity, linguistic diversity, and cultural context. The following are some of the future directions for AfricaNLP: 
% \begin{itemize}[leftmargin=3mm]
1) Explore Advanced NLP research topics for AfricaNLP, such as Text generation, reasoning, and vision-language tasks, which are nearly underexplored for African languages. 2) Using \texttt{AfricaNLPContributions} dataset, explore better modeling of automatic contribution classification tasks for NLP research papers. 3) Using the \textbf{AfricaNLP} paper collections, explore the papers in various ways, such as how venues influence the nature of AfricaNLP research over time. The opportunities we identified for AfricaNLP are presented in Appendix \ref{sec:opoprtunities}. %Continual tacking of the progress of AfricaNLP, using this reproducible work as an initial step.
% \end{itemize}
% Shared Tasks such as SemEval-2023 Task 12 \cite{muhammad2023afrisenti},  SemRel2024 \cite{ousidhoum-etal-2024-semrel2024}, and SemEval-2025 Task 11\cite{muhammad-etal-2025-semeval} have a statistically high number of papers that mention the task names and citations. Organizing such shared tasks will support the development of low-resource languages
\section{Conclusion}
In this work, we presented a systematic survey of AfricaNLP spanning the past two decades, examining contributions in terms of research outputs, authorship, institutional affiliations, and supporting organizations. Despite persistent challenges, interest in African NLP has expanded significantly, fueled by community initiatives and the development of new tools. Over this period, publications on African languages have more than quadrupled, reflecting the growing momentum of the field. Through a set of guiding research questions, we quantitatively captured this progress, highlighting community-driven collaborations in producing datasets and benchmarks.
% This work will serve as a foundational infrastructure layer, not only guiding funders and researchers but shaping inclusive AI policy and investment strategies. By identifying which languages and use cases are most underserved, the project will lay the groundwork for targeted future efforts in dataset development, model building, task development, and capacity strengthening across the continent. 
Our surveying pipeline is fully reproducible, extendable, and adaptable - enabling continuous tracking of AfricaNLP progress tools and offering a framework that can be applied to other languages and regions. 

\section*{Limitations}
The analysis of the AfricaNLP papers is done with the help of a series of automated preprocessing tools and LLMs for reproducibility. The processes may not be perfect, so some noise and errors may occur, starting from the scraping of available NLP papers. While Semantic Scholar accesses various research paper databases, some papers might be missed if the paper is not accessed by Semantic Scholar. This analysis aims to provide a higher-level view of what is happening in African NLP.
~\\ \textbf{Scope Limitations of Paper Searching:} Our paper searching scope encompasses the last 20 years, (2005 to 2025), based on the latest updates from Semantic Scholar and ACL databases. Additionally, we limited the scope of the study as follows.
\begin{itemize}[leftmargin=3mm]
    \item We survey papers that include at least one target keyword in their title or abstract; there might be interesting works that do not explicitly include an African country or language name in their title or abstract.
    \item Papers may not be included in the Semantic Scholar database, and papers might be published that are not accessed by Semantic Scholar and ACL anthology.
    \item Papers written only in English (the title and abstract) are included in the analysis.
    \item We exclude papers with very short abstracts (1/2 sentences), books, only linguistic papers without any computation, as our field of study is computer science OR linguistics, with/without other fields of study, based on the Semantic Scholar topic classifier models. 
\end{itemize}
\bibliography{anthology,custom}

\appendix
\onecolumn

\section*{Appendix}
\section{Processing of Affiliated Institutions and Acknowledged Organizations}
From 1,902 papers, the support contributors are identified in the acknowledgment section of 412 papers. The remaining papers are either do not have an acknowledgment section or acknowledge the support of research idea development, reviewers, readers, volunteers, or data annotators. During acknowledged organization annotation, similar organization names, such as Google DeepMind and Google Research are merged into a single organization (Google), together with support from different regional Google offices. The same approach is applied to acknowledged organizations: departments, faculties, and research labs or groups, which are grouped under the umbrella of the same organization names are recorded as the same affiliation. We extract the list of affiliated institutions from the actual papers. The affiliation institutions are identified from 1,367 papers; the other papers do not have an affiliation, either independent researchers, require a subscription to read/access the paper, or are unable ot access the PDF. 

\section{Additional Analysis}\label{app:analysis}
\paragraph{To what extent does this survey cover AfricaNLP research papers?}
To ensure that as many AfricaNLP papers as possible are included in our study, we manually collect a random 50 sample research papers from various publication venues such as \texttt{Springer Nature} and contributions (model, dataset, and methods) that involve African languages, and we cross-checked to ensure that these papers are included in our collection. Among these samples, 94\% of the papers are included in our collection. The AfricaNLP papers that do not have at least a single keyword in the title/abstract from the set of searching keywords might be excluded.

\paragraph{How effective are LLMs for paper contribution and NLP topic annotation?}
The papers are first annotated by LLMs and then verified/re-annotated by three human annotators, who are PhD students working in the NLP research area. We randomly sampled 500 papers (26\% of the collection) and annotated them using both LLMs and the three human annotators. We then measured the inter-annotator agreement (IAA) among pairs of LLMs, human annotators, and between humans and LLMs for each annotation task. The agreement between human annotators, as measured by Fleiss' Kappa (among the three annotators), is 84\%, while Cohen's kappa (pairwise agreement between human annotators, then averaged) is 86\%.

\paragraph{Which NLP topic(s) and languages received attention earlier in AfricaNLP?}
From the set of NLP research topics, 1) Syntax: tagging, chunking, and parsing \cite{fissaha-adafre-2005-part},  2) Phonology, morphology, and word segmentation \cite{hu-etal-2005-refining}, 3) ASR \cite{seid05_interspeech}, 4) OCR \cite{meshesha2007optical} are the early researched downstream NLP tasks since 2005. Languages such as Amharic, Afrikaans, Kinyarwanda, Swahili, Setswana, and Yoruba are among the earliest researched African languages since 2005. More details about the progress of specific NLP tasks are presented in Appendix \ref{app:tasks}.

\begin{figure*}[!h]
    \centering
    \includegraphics[width=\linewidth]{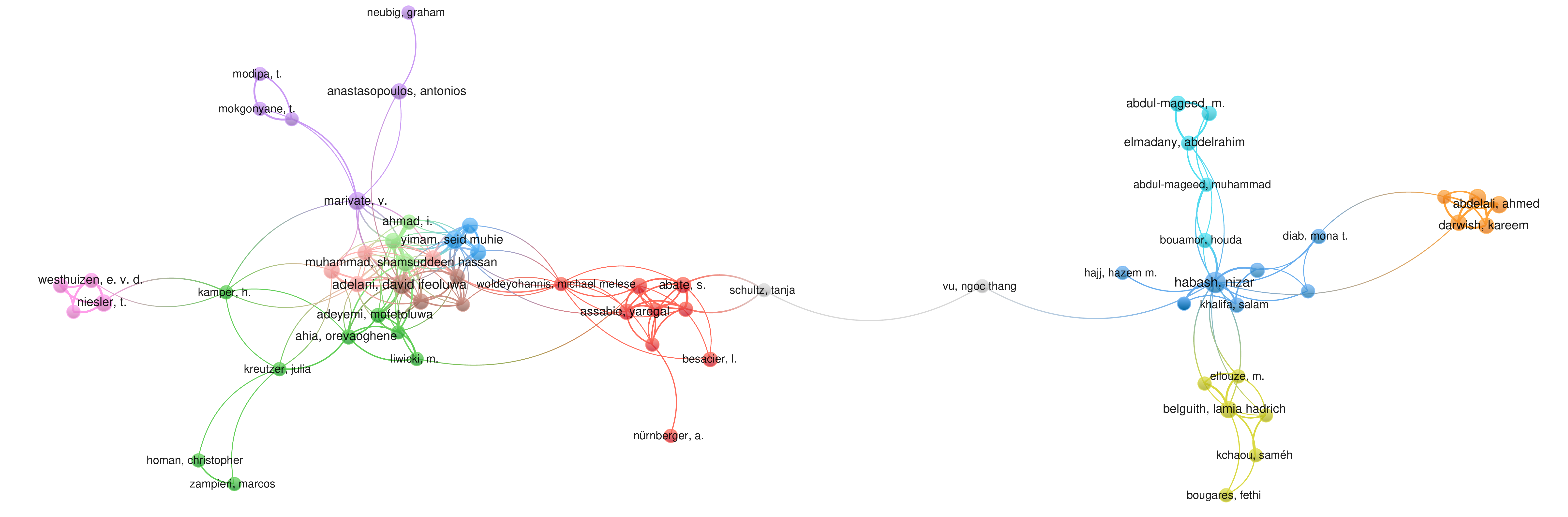}
    \caption{\textbf{Author co-authorship network analysis} with authors appearing in a minimum of 10 papers, 66 authors from 4901 authors, 12 clusters. The figure can be enlarged for better readability.}
    \label{fig:author-net-10}
\end{figure*}

\paragraph{Who are the authors contributing to AfricaNLP research?}
To examine author contributions and the distribution of papers, we conducted a statistical analysis of 1,902 publications. Parallel to the growth in publications, the number of contributing authors has increased exponentially over time. In 2006, there were 21 papers authored by 78 researchers. By 2024, this had grown to 287 papers with 1,103 authors. From 1,902 papers overall, there are 4,901 unique authors, with an average of 2.6 authors per paper. Among them, 52 authors have contributed more than 10 papers, 166 authors have contributed between 5 and 10 papers, and the remaining 1,683 authors have contributed between 1 and 5 papers. The authors and co-authorship networks are presented in Figure \ref{fig:author-net}, and authors who co-author in 10 or more papers are in Figure \ref{fig:author-net-10}.

\section{\texttt{AfricaNLPContributions} Dataset Distributions}
\begin{figure}[h!]
\centering

% Left side: image
\begin{minipage}[b]{0.45\textwidth}
    \centering
    \includegraphics[width=\linewidth]{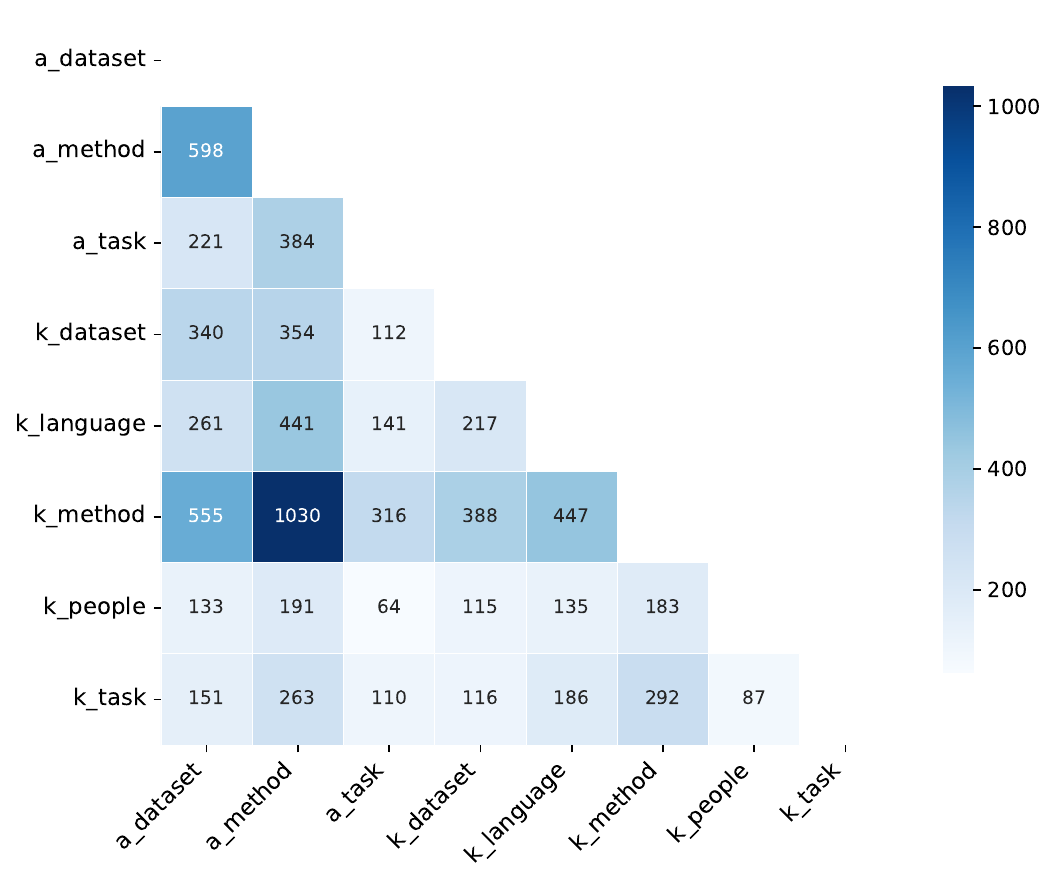}
    \caption{Contribution co-occurrence across the 8 contribution labels, 8,678 contribution sentences with 882 (11.3\%) statements received multiple contribution labels}
    \label{fig:occrance}
\end{minipage}
\hfill
% Right side: table
\begin{minipage}[b]{0.45\textwidth}
    \centering
% \begin{table}[t]
%     \centering
    % \scalebox{1.0}{
        \begin{tabular}{l l r r}
        \toprule
        {\bf Type} & {\bf Sub-typ.} & {\bf Prop. \%}& {\bf\# Sent} \\
        \midrule
        \multirow{5}{*}{Knowledge} & k-dataset & 8.16 & 708 \\
        & k-language & 10.73 & 931 \\
        & k-method & 26.98 & 2341 \\
        & k-people & 4.25 & 369 \\
        & k-task & 5.69 & 494 \\
        \midrule 
        \multirow{3}{*}{Artifact} & a-dataset & 12.93 & 1122 \\
        & a-method & 25.99 & 2255 \\
        & a-task & 5.28 & 458 \\
        \bottomrule
        \end{tabular}%}
    \caption{Distributions of \texttt{AfricaNLPContributions} classes: \textbf{Prop.} is proportion in percent and \textbf{Sent.} is number of sentences for the corresponding contribution class.}
    \label{tab:lbl_dist}
% \end{table}
    
\end{minipage}

\end{figure}

Figure \ref{fig:occrance} shows the co-occurrence of contribution types. Table \ref{tab:lbl_dist} shows the statistics of the 8 contribution types.

\section{Opportunities for AfricaNLP} \label{sec:opoprtunities}
To contribute more to AfricaNLP, we identified the following opportunities. %
\subsection{Community Engagement} We conducted a survey and compiled a list of communities and initiatives working on African AI/NLP. The following communities are frequently mentioned, such as Maskahene\footnote{\url{https://www.masakhane.io/}}, GhanaNLP\footnote{\url{https://ghananlp.github.io/}}, HuasaNLP\footnote{\url{https://hausanlp.github.io/}}, EthioNLP\footnote{\url{https://ethionlp.github.io/}}, NaijaVoices\footnote{\url{https://naijavoices.com/}}, Lanfrica\footnote{\url{https://lanfrica.com/en}}, and others are %Data science Africa, Deep learning Indaba
involving native speakers of African languages and Open for research collaborations including data collection, annotation, validation and model result error analysis ensures linguistic and cultural relevance in NLP systems.  The details about each community can be found on their website.
% Masakhane, EthioNLP, and HausaNLP are among the communities that are affiliated.

\subsection{Policy and Funding Support}
Increased funding and supportive policies have the potential to significantly strengthen NLP research in Africa by fostering innovations that address local needs. Supporting organizations play a central role in advancing research on African languages, and their involvement has expanded in parallel with the growth of publications. Specifically, the number of supporting organizations over the years was 26 in 2016, 14 in 2017, 28 in 2018, 39 in 2019, 64 in 2020, 63 in 2021, 103 in 2022, 146 in 2023, and 165 in 2024, corresponding to a total of 1,902 papers. %Notably, a single publication may be supported by one or multiple organizations, reflecting the collaborative nature of the research landscape. 
As shown in Figure \ref{fig:org}, the leading supporting organizations are predominantly based outside Africa, underscoring both the reliance on external institutions and the opportunities for these organizations to contribute to the development of low-resource languages. At the same time, the implementation of flexible AI policies at governmental and institutional levels \cite{siminyu2023consultative} is a critical pathway for advancing AfricaNLP.

\subsection{Science Assessment Reform and Inclusive Recognition}
Efforts to advance NLP for African languages must also engage with the structural enablers and barriers of scientific publishing. Many African institutions use rigid metrics that discourage conference participation, limiting visibility and uptake of local innovations in global NLP. Supporting reform of science assessment practices—to better align with the norms of fast-moving computational fields—is crucial. Advocacy is needed to: 1) Encourage recognition of top-tier NLP conference proceedings as equivalent to journal publications in research assessments; 2) Promote co-indexing of AfricaNLP outputs in both conference and journal repositories; 3) Create pathways for researchers to submit extended versions of ACL papers to indexed journals, increasing reward alignment; and 4) Support travel and registration sponsorships for African authors to present at global venues, ensuring that local innovations are seen, cited, and built upon. Such actions would not only increase participation but also elevate the global impact of African research communities.

\subsection{Limitations of Scientific Incentives and Dissemination Pathways}
A key limitation not captured in current bibliometric or semantic analyses is the influence of science incentive systems in different African countries. In many contexts, institutional and national research evaluation frameworks heavily prioritize journal publications—particularly those indexed in Scopus or Web of Science—over conference proceedings, even for fast-moving fields like NLP and machine learning. This may discourage African researchers from submitting to premier venues such as ACL, EMNLP, NAACL, or COLING, which are often conference-based, despite their high global impact. As a result, valuable African NLP work may remain underrepresented in global discourse or be siloed in lower-visibility outlets. ACL and other top NLP conferences might not reach most Africans who are in countries or institutions that reward journals. This publication misalignment contributes to systemic undercounting in automated surveys like ours, especially for papers published outside of major conference ecosystems. 
\section{LLM prompts for Annotation}\label{app:prompts}
% \section{Identifying NLPContributions from the papers}
% \section{Annotation Guidelines for human annotators}
% \begin{itemize}
%     \item at least containing one language in the NLP tasks
%     \item not only language, but also includes computations
% \end{itemize}

\subsection{LLM prompt to Identify AfricaNLP papers}
\begin{promptbox}[LLM prompts for Identify Africa NLP papers]
{\small
You are an assistant that determines whether a research paper focuses on Natural Language Processing (NLP) for African languages.\\
Based on the title and abstract below, determine if this paper is about NLP research involving African languages that develops NLP datasets, tools, models, techniques, or studies computational linguistics aspects of African languages.\\
Do not consider it if the title and abstract are not written in English.\\
Title: {title}\\
Abstract: {abstract}\\
Answer with only "yes" or "no" without any additional explanation.
}
\end{promptbox}

\subsection{LLM prompt to Annotate NLP Topics}
\begin{promptboxcn}[LLM prompt for annotating NLP topics]
{\small
You are an expert at analyzing NLP research papers. Based on the title and abstract below, extract the following information:\\
NLP TOPICS\\
Identify the specific NLP topic (category) addressed from the following list:\\
- \textit{\# List of NLP TOPICS that shown in Table \ref{tab:nlp_tasks} goes here ...}
\\
Output format:\\
\{
  "nlp-tasks": ["task1", "task2", ...]
\}
\\
Title: \{title\}\\
Abstract: \{abstract\}"""\\
Title: [PAPER-TITLE]\\
Abstract: [PAPER-ABSTRACT]\\
Answer:
}
\end{promptboxcn}

\subsection{LLM prompt to Annotate NLP Tasks}
\begin{promptbox}[LLM prompt for annotating NLP tasks]
{\small
You are an expert at analyzing NLP research papers. Based on the title and abstract below, extract the following information:\\
NLP TASKS\\
Identify the specific NLP tasks addressed from the following list:\\
- \textit{\# List of NLP TASKS goes here ...}
\\
Output format:\\
\{
  "nlp-tasks": ["task1", "task2", ...]
\}
\\
Title: \{title\}\\
Abstract: \{abstract\}"""\\
Title: [PAPER-TITLE]\\
Abstract: [PAPER-ABSTRACT]\\
Answer:
}
\end{promptbox}

\clearpage
\subsection{LLM prompt for NLP Paper Contribution Annotations}
\begin{promptbox}[LLM prompt for annotating abstracts for contribution and NLP tasks]
{\small
You are an expert at analyzing NLP research papers. Based on the title and abstract below, extract the following information:\\
\\
For each sentence in the abstract that describes a contribution, classify it into the appropriate category and extract the exact sentence. Group sentences by contribution type:\\
**artifact-method**: Introduces or proposes a new or novel NLP methodological approach or model to solve NLP tasks. [\textit{example here}]\\
**artifact-dataset**: Introduces a new NLP dataset (i.e., textual resources such as corpora or lexicons). [\textit{example here}]  \\
**artifact-task**: Introduces or proposes a new or novel NLP task formulation (i.e., well-defined NLP problem). [\textit{example here}]\\
**knowledge-method**: Describes new knowledge or insights about NLP models or methods. [\textit{example here}] \\
**knowledge-dataset**: Describes new knowledge about datasets, such as their new properties or characteristics. [\textit{example here}]\\
**knowledge-task**: Knowledge/insights about existing tasks or problem domains. [\textit{example here}]\\
**knowledge-language**: Presents new knowledge about language, such as a new property or characteristic of language. [\textit{example here}]\\
**knowledge-people**: Presents new knowledge about people, humankind, society, or communities. [\textit{example here}]\\
\\
OUTPUT FORMAT:\\
\\
\{\{
  "contribution": \{\{ "artifact-task": ["sentence 1", "sentence 2"],"artifact-method": ["sentence 1"], "artifact-dataset": [], "knowledge-method": ["sentence 1", "sentence 2"],"knowledge-dataset": ["sentence 1"], "knowledge-task": [], "knowledge-language": [], "knowledge-people": [] \}\},\\
  "nlp-tasks": ["task1", "task2", ...]\\
\}\}
\\
INSTRUCTIONS:\\
- A sentence might be assigned to more than one contribution category\\
- Multiple sentences can belong to the same category\\
Title: \{title\}\\
Abstract: \{abstract\}"""\\
Abstract: [PAPER-ABSTRACT]\\
Answer:\}
}
\end{promptbox}

\clearpage
\section{List of Support Contributor Organizations}\label{app:org}
Table \ref{fig:app-orgn} shows all the lists of organization names that support the papers.
\begin{table}[!h]
\centering
\resizebox{\textwidth}{!}{
\begin{tabular}{lclc}
\toprule
\textbf{Organization} & \textbf{\# Paper(s)} & \textbf{Organization} & \textbf{\# Paper(s)} \\
\hline
National Science Foundation & 52 & European Regional Development Fund & 2 \\
Google & 47 & Stanford University & 2 \\
National Research Foundation of South Africa & 42 & Howard University & 2 \\
Lacuna Fund & 25 & Carnegie Corporation of New York & 2 \\
OpenAI & 21 & Masaryk University & 2 \\
European Union & 19 & King Fahd University of Petroleum and Minerals & 2 \\
Canada Research Chairs & 17 & Czech Science Foundation & 2 \\
German Federal Ministry of Education and Research & 16 & Ministry of Education Youth and Sports of the Czech Republic & 2 \\
Deutsche Forschungsgemeinschaft & 14 & Allen Institute for Artificial Intelligence & 2 \\
Natural Sciences and Engineering Research Council of Canada & 14 & Federal Government of Nigeria & 2 \\
Social Sciences and Humanities Research Council of Canada & 13 & U.S. Army Research Office & 2 \\
Canadian Foundation for Innovation & 12 & Qatar Computing Research Institute & 2 \\
Science Foundation Ireland & 12 & Clarkson Open Source Institute & 2 \\
Digital Research Alliance of Canada & 11 & Swiss National Science Foundation & 2 \\
UBC Advanced Research Computing-Sockeye & 11 & Makerere University & 2 \\
Microsoft & 10 & University of Cambridge & 2 \\
NVIDIA & 10 & University of Oslo & 2 \\
European Research Council & 9 & Samsung & 2 \\
Intelligence Advanced Research Projects Activity & 9 & Research Council of Norway & 2 \\
Defense Advanced Research Projects Activity & 9 & University of Malawi & 2 \\
University of Cape Town & 8 & Covenant University & 2 \\
Qatar Foundation & 8 & University of Texas & 2 \\
French Research Agency & 7 & Indiana University & 2 \\
University of Hamburg & 7 & University of Sheffield & 2 \\
New York University Abu Dhabi & 7 & Qatar National Research Fund & 2 \\
Meta & 6 & American University of Beirut & 2 \\
Amazon & 6 & Birzeit University & 2 \\
Bill \& Melinda Gates Foundation & 6 & National Endowment for the Humanities & 2 \\
Oracle & 5 & Israeli Science Foundation & 2 \\
University of Bremen & 5 & German Ministry of Education and Science & 2 \\
Johns Hopkins University & 5 & National Key Research and Development Program of China & 2 \\
Masakhane & 5 & Princeton Laboratory for Artificial Intelligence & 2 \\
University of Gothenburg & 5 & Nigerian Artificial Intelligence Research Scheme & 1 \\
South African Centre for Digital Language Resources & 5 & Label Studio & 1 \\
Swedish Research Council & 5 & MUR FARE 2020 & 1 \\
Advanced Micro Devices & 5 & Trelis Research & 1 \\
University of Edinburgh & 4 & UK Research and Innovation & 1 \\
African Institute for Mathematical Sciences & 4 & Luxembourg Ministry of Foreign and European Affairs & 1 \\
Foundation for Science and Technology, PT & 4 & CONAHCYT & 1 \\
South African Department of Arts and Culture & 4 & AI for Good & 1 \\
Intron Health & 4 & TUM School of Social Sciences and Technology & 1 \\
University of Porto & 4 & Hasso Plattner Institute for Digital Engineering & 1 \\
Addis Ababa University & 4 & CNPq,FAPEMIG,FAPESP, Brazil & 1 \\
German Academic Exchange Service & 4 & University of Information Technology,Vietnam & 1\\
National Natural Science Foundation of China & 4 & Alberta Innovates & 1 \\
Stony Brook University & 3 & Ministry of Science Technology and Innovation, Brazil & 1 \\
University of Oregon & 3 & Stanford UIT & 1 \\
Toloka & 3 & ABSA Data Science & 1 \\
Huawei & 3 & KAIST-NAVER AI Center & 1 \\
Ghent University & 3 & Korea Government MSIT & 1 \\
TensorFlow & 3 & GENCI-IDRIS & 1 \\
Jimma University & 3 & ESPERANTO & 1 \\
George Mason University & 3 & Helmholtz Programme-oriented Funding & 1 \\
Ashesi University & 3 & GRAIN & 1 \\
Bahir Dar University & 3 & Science Foundation Ireland Digital Reality & 1 \\
Japan Society for the Promotion of Science & 3 & Canada CIFAR AI Chair & 1 \\
Waterloo AI Institute & 3 & German Research Center for Artificial Intelligence & 1 \\
Microsoft Research & 3 & University of Hull & 1 \\
South African National Research Foundation & 3 & TRADEF & 1 \\
U.S. Army Research Laboratory & 3 & Stony Brook Research Computing & 1 \\
Mila - Quebec AI Institute & 3 & University of Murcia & 1 \\
National Research Foundation of Korea & 3 & &  \\
Compute Canada & 3 & &  \\
\bottomrule

\end{tabular}
}
\caption{List of organizations with the number of papers mentioned by}
\label{fig:app-orgn}

\end{table}

\section{Partial List of Affiliation Organizations}\label{app:intitutions}
Table \ref{fig:app-inst} shows all the lists of institution names that support the papers.
\begin{table}[!h]
\centering
\resizebox{\textwidth}{!}{
\begin{tabular}{lclc}
\toprule
\textbf{Institutions} & \textbf{\# Paper(s)} & \textbf{Organization} & \textbf{\# Paper(s)} \\
\hline
University of Pretoria & 53 & University of the Witwatersrand & 10 \\
Carnegie Mellon University & 44 & Nnamdi Azikiwe University & 10 \\
Saarland University & 38 & Dublin City University & 10 \\
MBZUAI & 33 & Mila Quebec AI Institute & 10 \\
Stellenbosch University & 33 & University of Waterloo & 10 \\
Masakhane & 31 & University of Stuttgart & 10 \\
University of Cape Town & 27 & Qatar University & 10 \\
Addis Ababa University & 27 & University of Maryland & 9 \\
Columbia University & 25 & University of Sheffield & 9 \\
Johns Hopkins University & 23 & Massachusetts Institute of Technology & 9 \\
North-West University & 23 & University of Illinois & 9 \\
Bahir Dar University & 22 & Georgetown University & 9 \\
New York University & 22 & Kaduna State University & 9 \\
Qatar Computing Research Institute & 22 & LMU Munich & 9 \\
The University of British Columbia & 21 & University of Ibadan & 9 \\
University College London & 20 & African Institute for Mathematical Sciences & 9 \\
University of Edinburgh & 20 & Stony Brook University & 9 \\
New York University Abu Dhabi & 19 & University of Porto & 9 \\
McGill University & 19 & University of Bergen & 9 \\
University of Washington & 19 & University of Zimbabwe & 9 \\
Imperial College London & 18 & Cairo University & 9 \\
Google & 18 & Bayero University Kano & 8 \\
Uppsala University & 17 & Cardiff University & 8 \\
Makerere University & 17 & Montclair State University & 8 \\
Technical University of Munich & 16 & University of British Columbia & 8 \\
Indiana University & 16 & American University of Beirut & 8 \\
University of Cambridge & 16 & Instituto Politécnico Nacional & 7 \\
University of Pennsylvania & 16 & University of Minnesota & 7 \\
University of Sfax & 16 & Amazon & 7 \\
Stanford University & 15 & Luleå University of Technology & 7 \\
Lelapa AI & 14 & Invertible AI & 7 \\
George Mason University & 14 & University of Oslo & 7 \\
Lancaster University & 13 & Idiap Research Institute & 7 \\
University of Hamburg & 13 & University of Southern California & 7 \\
University of Leeds & 13 & The George Washington University & 7 \\
Alexandria University & 13 & University of Nairobi & 7 \\
Hamad Bin Khalifa University & 13 & Arba Minch University & 7 \\
University of Gothenburg & 13 & University of the Western Cape & 7 \\
University of South Africa & 13 & Maseno University & 6 \\
Northeastern University & 12 & SADiLaR & 6 \\
Universität Hamburg & 12 & University of Amsterdam & 6 \\
University of California & 12 & Avignon University & 6 \\
Ahmadu Bello University & 12 & Meta & 6 \\
Bayero University & 12 & Inria & 6 \\
HausaNLP & 11 & Clarkson University & 6 \\
Wollo University & 11 & University of Melbourne & 6 \\
Rochester Institute of Technology & 11 & University of Helsinki & 6 \\
Microsoft & 11 & University of Limpopo & 6 \\
Ghent University & 11 & LIMSI-CNRS & 6 \\
Obafemi Awolowo University & 11 & TshwaneDJe HLT & 6 \\
University of Toronto & 10 & University of Birmingham & 6 \\

\bottomrule
\end{tabular}
}
\caption{List of institutions/organizations affiliated with AfricaNLP papers with the number of papers mentioned.  There are other more than 100 affiliations mentioned less than 6 times.}
\label{fig:app-inst}
\end{table}

\clearpage
\section{Full List of NLP Topics} \label{app:tasks}
Table \ref{tab:nlp_tasks} shows all the lists of NLP topics with the number of papers.
\begin{table}[!h]
\centering
\resizebox{\textwidth}{!}{
\begin{tabular}{lclc}
\toprule
\textbf{Task name} & \textbf{\# Paper(s)} & \textbf{Task name} & \textbf{\# Paper(s)} \\
\hline
Low-resource language processing & 896 & Text generation & 23 \\
Multilinguality and language diversity & 638 & Educational NLP & 22 \\
Resources and evaluation & 541 & Conversational AI and chatbots & 21 \\
Phonology, morphology and word segmentation & 291 & Natural language understanding & 20 \\
Machine translation & 276 & Discourse and pragmatics & 19 \\
Speech recognition, text-to-speech and spoken language understanding & 273 & Few-shot and zero-shot learning & 17 \\
Machine learning for NLP & 247 & Sign language processing & 16 \\
Sentiment analysis, stylistic analysis, and argument mining & 209 & Benchmarking and testing & 15 \\
Syntax: tagging, chunking and parsing & 145 & Computational social science and cultural analytics & 13 \\
Dialect and language identification & 108 & Interpretability and analysis of models for NLP & 12 \\
Representation learning & 102 & Computational linguistics theory & 11 \\
Semantics: lexical and sentence-level & 89 & Human-centered NLP & 10 \\
Transfer learning and domain adaptation & 86 & Legal and biomedical NLP & 9 \\
Code-switching and multilingual processing & 75 & Text summarization & 8 \\
Information retrieval and text mining & 67 & Dialogue and interactive systems & 8 \\
Language modeling & 60 & Reasoning & 6 \\
Information extraction & 59 & Knowledge graph construction and reasoning & 4 \\
Social media and web text analysis & 57 & Privacy and security in NLP & 4 \\
Linguistic theories, cognitive modeling and psycholinguistics & 53 & Vision-language tasks & 4 \\
Efficient/low-resource methods for NLP & 51 & Document understanding and layout analysis & 2 \\
Ethics, bias, and fairness & 36 & Creative writing and storytelling & 2 \\
Multimodality and language grounding to vision, robotics and beyond & 30 & Text-to-code generation & 1 \\
NLP applications & 28 & Paraphrasing and text rewriting & 1 \\
Error correction & 25 & Personality and psychological profiling from text & 1 \\
Question answering & 24 & Temporal information processing & 1 \\
\hline
\end{tabular}
}
\caption{NLP task distribution by number of publications/papers}
\label{tab:nlp_tasks}
\end{table}

\section{NLP tasks and progresses} \label{app:nlptopics-tasks}
\begin{figure*}[h!]
    \centering
    \includegraphics[width=\linewidth]{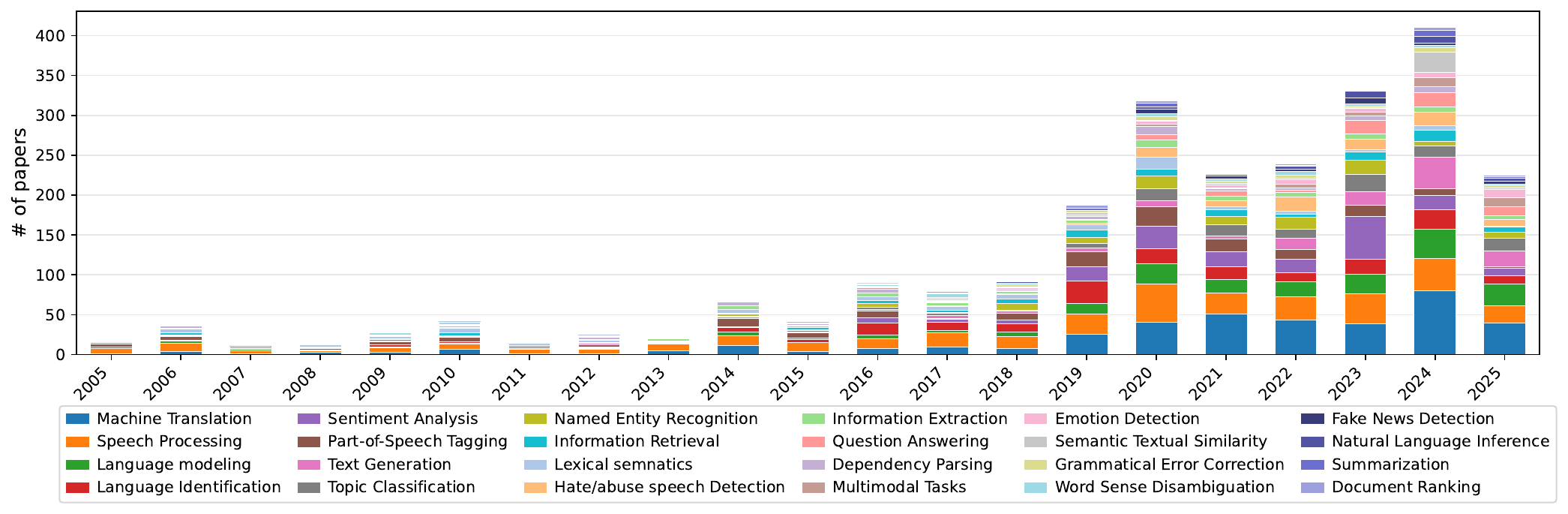}
    \caption{AfricaNLP Downstream NLP task progresses categories,  (a paper may appear in multiple categories). The NLP topics are sorted in descending order based on the overall total frequency of papers.}
    \label{fig:tasks2}
\end{figure*}

Table \ref{fig:tasks2} shows lists of NLP tasks progress across years.

\clearpage
\section{Human Annotation Interface}
\begin{figure*}[!h]
  \centering
  \includegraphics[width=0.48\linewidth]{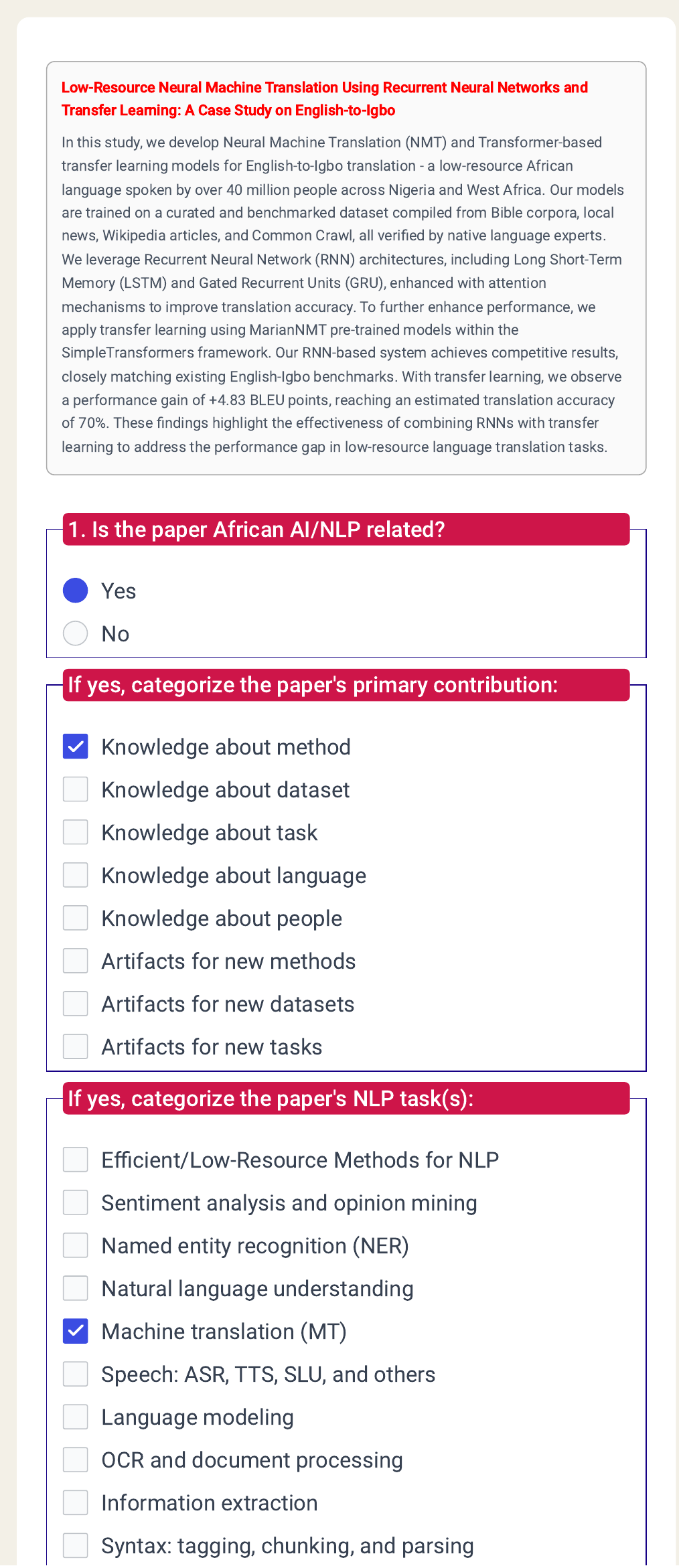} \hfill
  \includegraphics[width=0.48\linewidth]{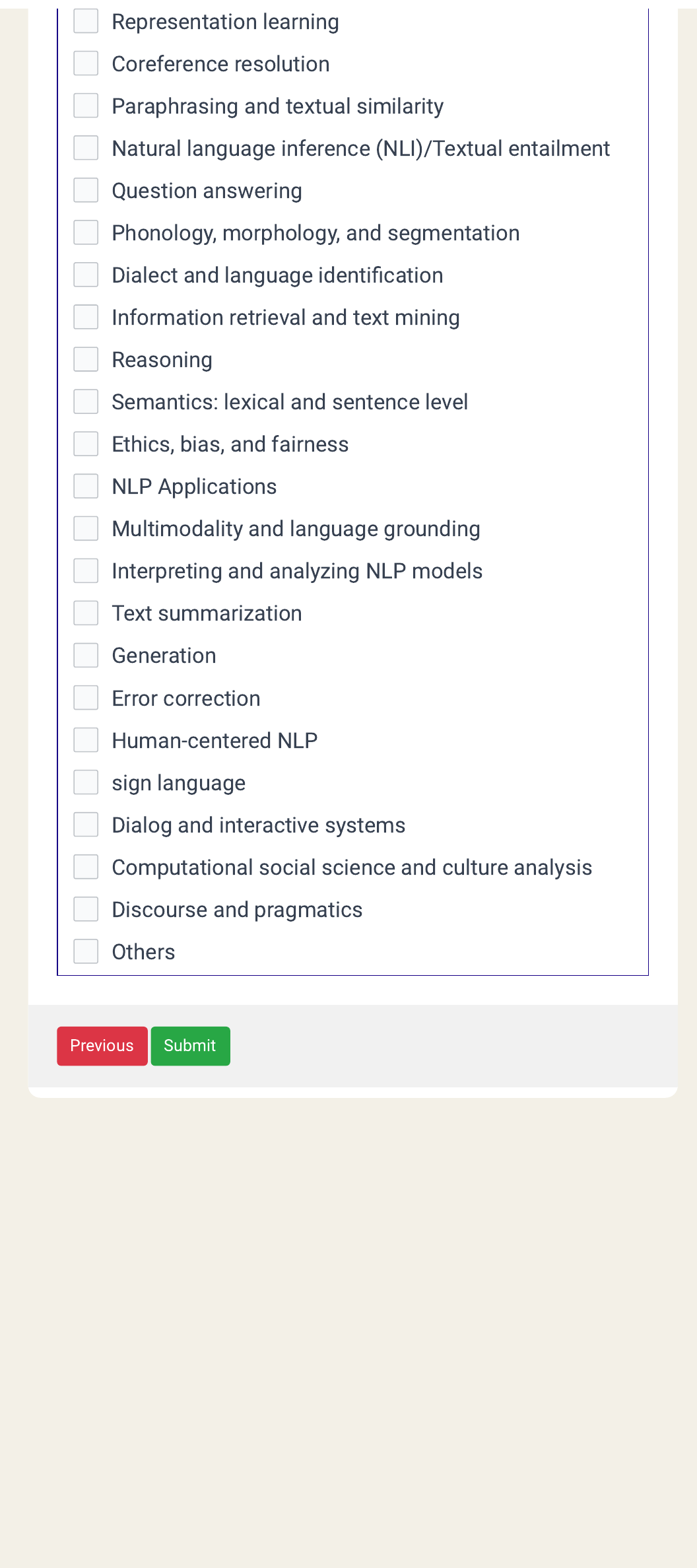}
  \caption {NLP paper annotation tool interface screenshot, customized version of POTATO annotation tool \cite{pei-etal-2022-potato}.}
  \label{fig:ui}
\end{figure*}

Figure \ref{fig:ui} shows the human annotation interface for identifying AfricaNLP papers, contributions, and NLP topic annotations

\section{Contribution Classifier Training Details}\label{app:param}
\begin{itemize}
    \item The PLMs are loaded from the Transformer on Huggingface
    \item BERT = google-bert/bert-base-uncased
    \item SciBERT = allenai/scibert\_scivocab\_uncased
    \item AfroXLMR = Davlan/afro-xlmr-large-76L
    \item GPT\footnote{\url{https://openai.com/index/gpt-4-1/}}  is Gpt-4.1-mini-2025-04-14 \cite{gpt4.1} and Gemini-2.5-flash \cite{gemini2.5}
    \item Traing details for all PLMs are batch size = 16, epoch = 3, zero shot for LLMs
\end{itemize}

%The top frequently used keywords from countries and languages are shown in Figure \ref{fig:keywords}. The reasons for most of the papers are not African NLP papers are: 1) written in languages other than English, linguistic papers without any computation, and not NLP papers. 

% \begin{figure}[!h]
%     \centering
%     \includegraphics[width=\linewidth]{images/country-language.pdf}
%     \caption{Top occurrences of country and language keywords}
%     \label{fig:keywords}
% \end{figure}
\section{Affiliation institutions networks of AfricaNLP }
\begin{figure*}[!h]
    \centering
    \includegraphics[width=\linewidth]{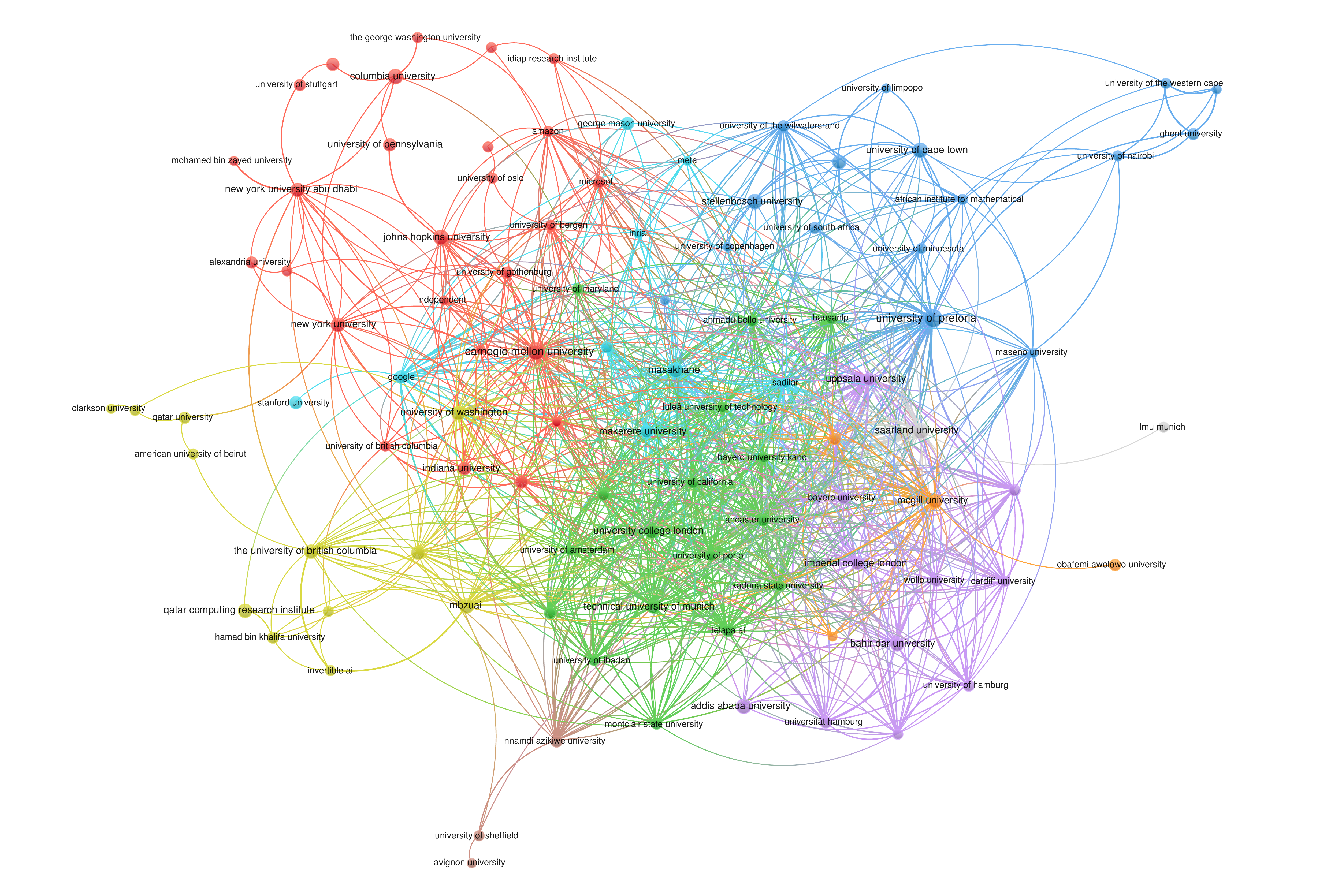}
    \caption{\textbf{Affiliation networks}: The networks are generated using VOSviewer with a minimum of five documents per author for better readability. The size of the nodes (circles) and edges (lines) corresponds to the frequency of affiliations and links, respectively. In the network visualization, each node represents an affiliation, and a cluster with the same color indicates affiliations within a close network or collaboration. Connected authors account for 123 out of more than 200 affiliations, grouped into 10 clusters. The figure can be enlarged for better readability.}
    \label{fig:affilation-net}
\end{figure*}

Figure \ref{fig:affilation-net} shows the networks of Affiliation institutions, organizations, and individual researchers from AfricaNLP papers. The visualization is done with the VOSviewer\footnote{\url{https://www.vosviewer.com/}} bibliometric analysis tool.

\newpage
\section{Author Networks of AfricaNLP Papers}
\begin{figure*}[!h]
    \centering
    \includegraphics[width=\linewidth]{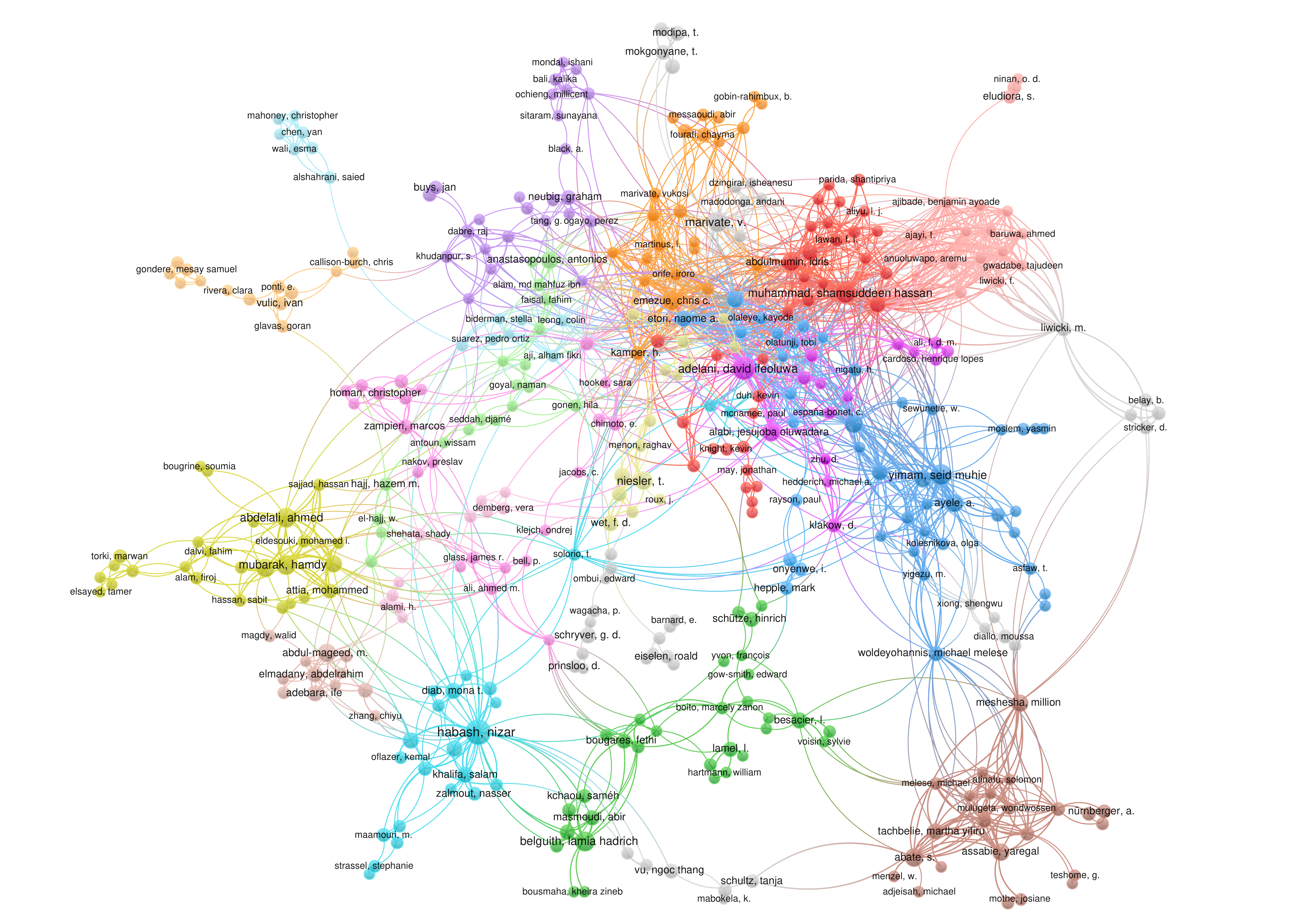}
    \caption{\textbf{Author co-authorship network analysis} generated using VOSviewer with a minimum number of papers of an author equal to five for better readability. Connected authors comprise 173 out of 4,901 authors, grouped into 6 clusters. In the network visualization, each node represents an author, and a cluster with the same color indicates authors with a close relationship. The size of the node shows the frequency of occurrence contributing to research in the past 20 years. The figure can be enlarged for better readability.}
    \label{fig:author-net}
\end{figure*}

Figure \ref{fig:author-net} shows the networks of authors' co-authorship from AfricaNLP papers.

\newpage
\section{Search Keyword Networks}
\begin{figure*}[!h]
    \centering
    \includegraphics[width=\linewidth]{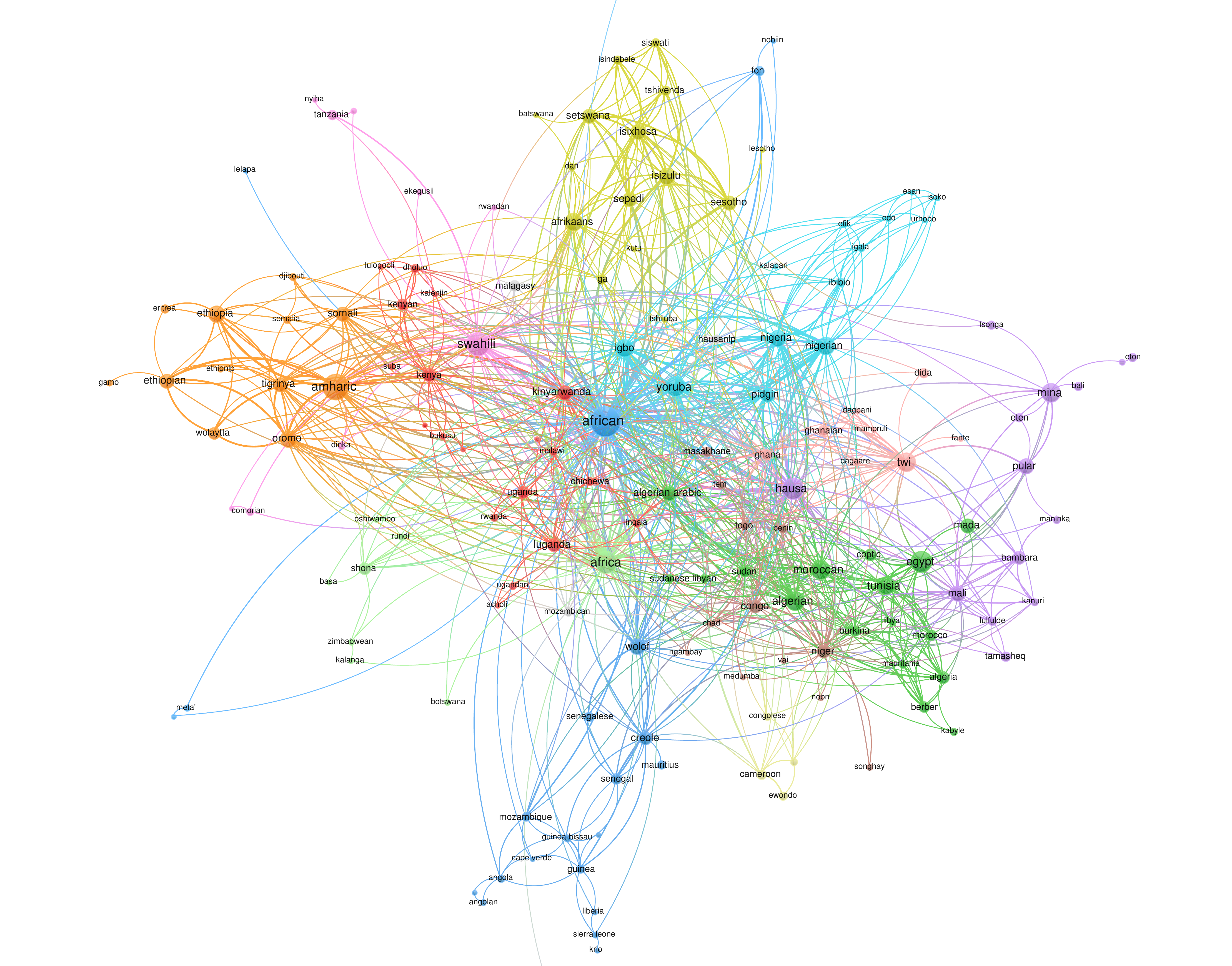}
    \caption{\textbf{Keyword co-occurrence network analysis} generated using VOSviewer with a minimum occurrence of five for better readability. Connected keywords comprise 121 out of 184 keywords, grouped into 11 clusters. In the network visualization, each node represents a keyword, and a cluster with the same color indicates keywords they are appear together in the paper. The size of the node indicates the frequency of occurrence in the AfricaNLP papers. The figure can be enlarged for better readability.}
    \label{fig:keyword-net}
\end{figure*}

Figure \ref{fig:keyword-net} shows the networks of keyword co-occurrence from AfricaNLP papers. Out of the set of more than 600 country names (including adjectival forms like African), language names, and initiative organization community names, only 184 keywords have at least one paper.

\newpage
\section{Examples of Paper annotation disagreement between LLMs}
Table \ref{tab:llm-disargeement} shows examples of papers that the LLMs disagree on when classifying as AfricaNLP NLP papers.

\begin{table*}[!th]
    \centering
    \small
    \begin{adjustbox}{width=\columnwidth}
        \begin{tabular}{p{4cm} p{11.0cm} >{\centering\arraybackslash}p{1cm} >{\centering\arraybackslash}p{1cm}}

        \toprule
        \textbf{Paper} & \textbf{Title} & \textbf{GPT} & \textbf{Gemini}\\
        \midrule
        \citet{hossain-etal-2025-enhancing} & Enhancing Dialectal Arabic Intent Detection through Cross-Dialect Multilingual Input Augmentation &\ding{55} & \ding{51} \\
        \citet{khered-etal-2025-dial2msa} & Dial2MSA-Verified: A Multi-Dialect Arabic Social Media Dataset for Neural Machine Translation to Modern Standard Arabic &\ding{55} & \ding{51} \\
        \citet{miyagawa-2025-rag}&RAG-Enhanced Neural Machine Translation of Ancient Egyptian Text: A Case Study of THOTH AI&\ding{55} & \ding{51} \\
        
        \citet{malaysha-etal-2024-nlu}&NLU-STR at SemEval-2024 Task 1: Generative-based Augmentation and Encoder-based Scoring for Semantic Textual Relatedness&\ding{55} & \ding{51} \\
        \citet{alwajih-etal-2024-dallah}&Dallah: A Dialect-Aware Multimodal Large Language Model for Arabic&\ding{55} & \ding{51} \\
        \cite{elfqih-monti-2024-large}&Large Language Models as Legal Translators of Arabic Legislation: Do ChatGPT and Gemini Care for Context and Terminology?&\ding{55} & \ding{51} \\
        
        \citet{hadj-mohamed-etal-2023-alphamwe}&AlphaMWE-Arabic: Arabic Edition of Multilingual Parallel Corpora with Multiword Expression Annotations&\ding{55} & \ding{51} \\
        \cite{sahyoun-shehata-2023-aradiawer}&AraDiaWER: An Explainable Metric For Dialectical Arabic ASR&\ding{55} & \ding{51} \\

        \citet{tirosh-becker-etal-2022-part}&Part-of-Speech and Morphological Tagging of Algerian Judeo-Arabic&\ding{55} & \ding{51} \\
        \cite{hassib-etal-2022-aradepsu}&AraDepSu: Detecting Depression and Suicidal Ideation in Arabic Tweets Using Transformers&\ding{55} & \ding{51} \\
        
        \citet{mubarak-etal-2021-arabic}&Arabic Offensive Language on Twitter: Analysis and Experiments&\ding{55} & \ding{51} \\
        
        \citet{10.1145/3322905.3322931}&Optical Character Recognition for Coptic fonts: A multi-source approach for scholarly editions&\ding{55} & \ding{51} \\

        \citet{feleke-2017-similarity}&The similarity and Mutual Intelligibility between Amharic and Tigrigna Varieties&\ding{55} & \ding{51} \\
        \citet{aftiss-etal-2025-empirical}&Empirical Evaluation of Pre-trained Language Models for Summarizing Moroccan Darija News Articles&\ding{51}&\ding{55}\\
        \citet{sang-etal-2025-federated}&Federated Meta-Learning for Low-Resource Translation of Kirundi &\ding{51}&\ding{55}\\
        \citet{tashu-tudor-2025-mapping}&Mapping Cross-Lingual Sentence Representations for Low-Resource Language Pairs Using Pre-trained Language Models&\ding{51}&\ding{55}\\
        \citet{bamutura-etal-2020-towards}&Towards Computational Resource Grammars for Runyankore and Rukiga&\ding{51}&\ding{55}\\
        \citet{yuan-etal-2020-interactive}&Interactive Refinement of Cross-Lingual Word Embeddings&\ding{51}&\ding{55}\\

        \bottomrule
        \end{tabular}
    \end{adjustbox}
    % }
    \caption{LLMs answer for the question \textit{Is this African AI/NLP paper?} using the title and abstract of the paper. Note that the prompt includes an abstract of the paper. For simplicity, we provide here the title of the paper only. \ding{55} is not AfricaNLP paper response and \ding{51} is Yes reprocess. These papers are part of our collection, which is handled by human evaluations after the LLM predictions.}
    \label{tab:llm-disargeement}
    % \vspace*{-4mm}
\end{table*}

\newpage
\section{Number of publication across languages versus number of (both L1 and L2) speakers of the language}
 \textbf{How does the number of language speakers correlate with the availability of language resources/publications?}. The 17 top languages, along with their number of speakers and publications from 1902 papers, are presented in Table \ref{tab:speaker}. The analysis reveals a weak correlation between the number of speakers and research attention across African languages. While Swahili and Nigerian Pidgin have the largest speaker bases (150M and 121M, respectively), they are comparatively underrepresented in research publications, particularly Nigerian Pidgin with only 25 papers. In contrast, Amharic, with 57M speakers, shows a disproportionately high research presence (204 papers), suggesting that academic focus and resource availability outweigh population size in influencing research activity. East African languages such as Amharic and Swahili receive more attention overall compared to West African and Southern African languages, despite the latter regions hosting several high-population languages like Hausa and Igbo. Interestingly, languages with relatively smaller populations, such as Twi and Afrikaans, also demonstrate notable research engagement, underscoring the role of institutional support, data accessibility, and linguistic policy rather than speaker population in driving research output. There is little to no direct correlation between the number of language speakers and the availability of language resources or publications, as research attention is more strongly influenced by institutional support, data availability, and linguistic policy than by speaker population size.
\begin{table}[!h]
    \centering
    \resizebox{\columnwidth}{!}{
    \begin{tabular}{llcc}
    \toprule
       \textbf{Language name} & \textbf{Region (of Africa)} & \textbf{\# of Speakers} & \textbf{\# of papers} $\downarrow$ \\
       \hline
       Amharic (amh) &East & 57M & 204\\
       Swahili (swa) &East/Central & 150M & 140 \\
       Twi (twi)     &West & 9M & 105 \\
       Yoruba (yor)  &West & 46M & 84 \\
       Hausa (hau)   &West &77M & 73 \\
       Afrikaans (afr)&Southern  &7M & 47\\
       Oromo (orm)    &East  & 37M & 42\\
       Isizulu (zul)  &Southern  & 27M&40\\
       Igbo (ibo)     &West  & 31M& 39\\
       Wolof (wol)    &West  &5M & 32\\
       Kinyarwanda (kin)&East  & 10M& 29\\
       Sesotho (sot)  &Southern & 6M & 28\\
       Isixhosa (xho) &Southern &19M& 28\\
       Nigerian Pidgin (pcm)&West  &121M&25\\
       Sepedi (nso) &Northeastern &6M&25\\
       Tigrinya (tir) &East  &9M&24\\
       Setswana (tsn) &Southern  &14M&24\\
       \bottomrule
    \end{tabular}}
    \caption{Languages number of L1 and L2 speakers vs number papers that has > 20 papers. Note that this frequency reflects the number of language keyword occurrences in the titles and abstracts of papers; some papers may involve a language without explicitly mentioning its name in the title or abstract. Number speakers are sourced from \citet{ojo-etal-2025-afrobench}.}
    \label{tab:speaker}
\end{table}

\newpage
\section{AfricaNLP Research Explorer Tool}\label{app:tool}
\begin{figure*}[!h]
    \centering
    \includegraphics[width=\linewidth]{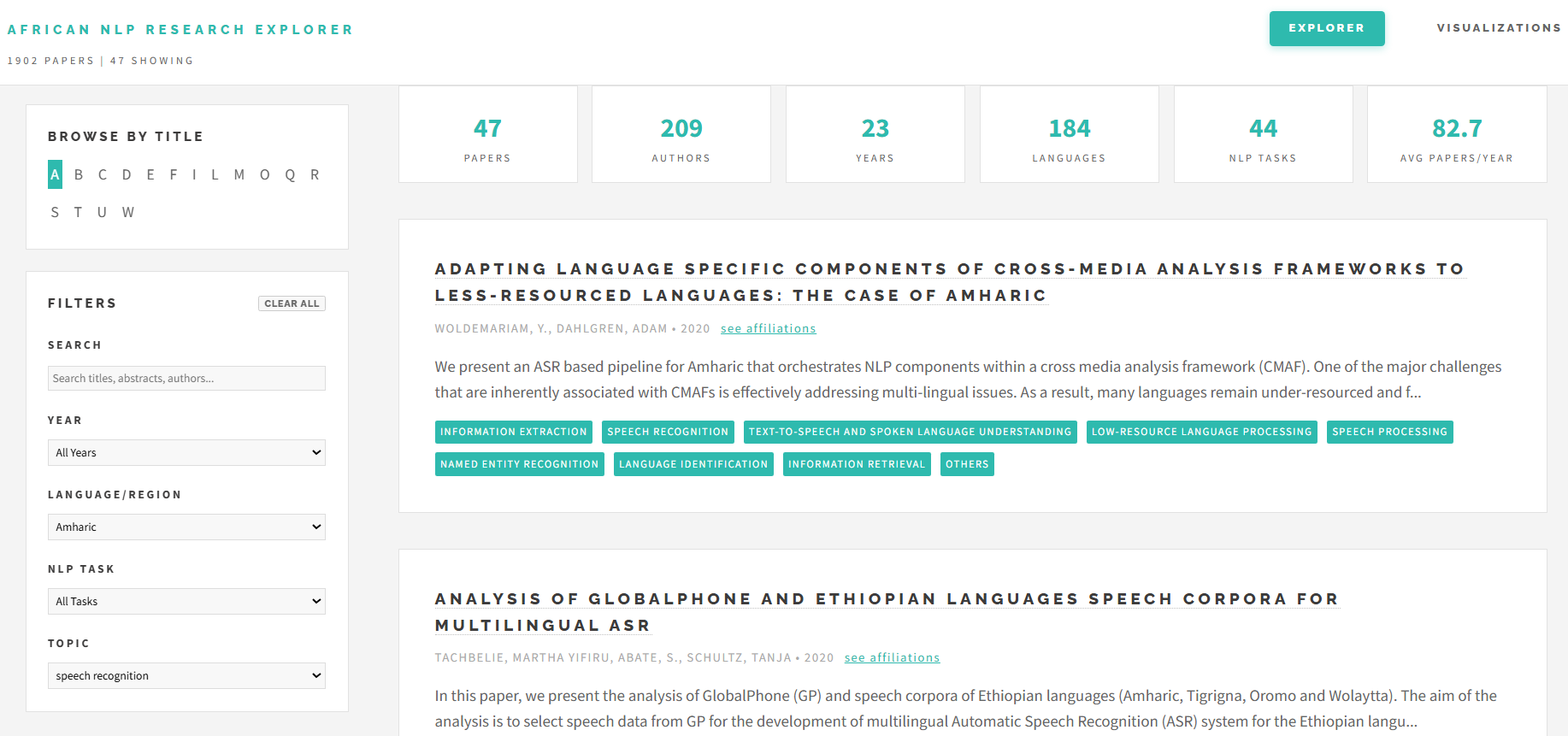}
    \caption{AfricaNLP research explorer tool screenshot (under improvement) and for Amharic language - speech recognition papers as a demonstration} %AfricaNLP workshop co-located with ICLR 2020 started and has published a range of papers, including research on machine translation for 38 African languages, authored by over 35 contributors \citet{orife2020masakhanemachinetranslation}.}
    \label{fig:tool}
\end{figure*}

Figure \ref{fig:tool} shows the underconstruction AfricaNLP research explorer tool. It features paper searching capabilities, including the ability to search for paper titles and abstracts, as well as categories by language, NLP topic, and NLP task. The tool link will be provided during the camera-ready submission of the paper.

\end{document}